%% file: ijcai24.tex
\documentclass{article}
\usepackage{ijcai24}

\usepackage[linesnumbered,ruled,lined]{algorithm2e}
\usepackage{multirow}
\usepackage{enumitem}
\usepackage{times}
\usepackage{soul}
\usepackage{url}
\usepackage[hidelinks]{hyperref}
\usepackage[utf8]{inputenc}
\usepackage[small]{caption}
\usepackage{graphicx}
\usepackage{amsmath}
\usepackage{amsthm}
\usepackage{booktabs}
\usepackage[switch]{lineno}

\usepackage{booktabs}
\usepackage{subcaption}
\usepackage{times}
\usepackage{epsfig}

\usepackage{graphicx}
\usepackage{amsmath}
\usepackage{amssymb}
\usepackage{booktabs}

\usepackage{xcolor,pifont}
\newcommand*\colourcheck[1]{%
  \expandafter\newcommand\csname #1check\endcsname{\textcolor{#1}{\ding{52}}}%
}
\colourcheck{blue}
\colourcheck{green}
\colourcheck{red}
\usepackage{pifont}
\newcommand{\xmark}{\ding{55}}%

\title{A Self-explaining Neural Architecture for Generalizable Concept Learning}

\author{
Sanchit Sinha$^1$
\and
Guangzhi Xiong$^1$\And
Aidong Zhang$^{1}$\\
\affiliations
$^1$University of Virginia\\
\emails
\{sanchit, hhu4zu, aidong\}@virginia.edu
}

\begin{document}

\maketitle

\begin{abstract}
With the wide proliferation of Deep Neural Networks in high-stake applications, there is a growing demand for explainability behind their decision-making process. Concept learning models attempt to learn high-level `concepts' - abstract entities that align with human understanding, and thus provide interpretability to DNN architectures. However, in this paper, we demonstrate that present SOTA concept learning approaches suffer from two major problems - lack of \textbf{concept fidelity} wherein the models fail to learn consistent concepts among similar classes and limited \textbf{concept interoperability} wherein the models fail to generalize learned concepts to new domains for the same task. Keeping these in mind, we propose a novel self-explaining architecture for concept learning across domains which - i) incorporates a new \textit{concept saliency network} for representative concept selection, ii) utilizes \textit{contrastive learning} to capture representative domain invariant concepts, and iii) uses a novel \textit{prototype-based concept grounding} regularization to improve concept alignment across domains. We demonstrate the efficacy of our proposed approach over current SOTA concept learning approaches on four widely used real-world datasets. Empirical results show that our method improves both concept fidelity measured through concept overlap and concept interoperability measured through domain adaptation performance. 

\end{abstract}

\input{1_introduction}
\input{2_related}
\input{3_method}

\input{4_experiments}

\input{6_conclusion}






\bibliographystyle{named}
\bibliography{ref-short}

\appendix
\input{appendix}

\end{document}

%% file: 1_introduction.tex
\section{Introduction}
\label{sec:intro}

Deep Neural Networks (DNNs) have revolutionized a variety of human endeavors from vision to language domains. Increasingly complex architectures provide state-of-the-art performance which, in some cases has surpassed even human-level performance. Even though these methods have incredible potential in saving valuable man-hours and minimizing inadvertent human mistakes, their adoption has been met with rightful skepticism and extreme circumspection in critical applications like medical diagnosis \cite{liu2021advances,aggarwal2021diagnostic}, credit risk analysis \cite{szepannek2021facing}, etc. 

With the recent surge in interest in Artificial General Intelligence (AGI) through DNNs, the broad discussion around the lack of rationale behind DNN predictions and their opaque decision-making process has made them notoriously \textbf{black-box} in nature \cite{rudin2019stop,varoquaux2022machine,d2020underspecification,weller2019transparency}. In extreme cases, this can lead to a lack of \textit{alignment} between the designer's intended behavior and the model's actual performance. For example, a model designed to analyze and predict creditworthiness might look at features that should not play a role in the decision such as race or gender \cite{bracke2019machine}. This, in turn, reduces the trustworthiness and reliability of model predictions (even if they are correct) which defeats the purpose of their usage in critical applications \cite{hutchinson201950,raji2020closing}.

In an ideal world, DNNs would be inherently explainable by their \textit{inductive biases}, as it is designed keeping stakeholders in account. However, such an expectation is gradually relaxed with the increasing complexity of the data which in itself drives up the complexity of the architectures of DNNs to fit said data. Several approaches to interpreting DNNs have been proposed. Some approaches assign relative importance scores to features deemed important like LIME \cite{ribeiro2016should}, Integrated Gradients \cite{sundararajan2017axiomatic}, etc. Other approaches rank training samples by their importance to prediction like influence functions \cite{koh2017understanding}, data shapley \cite{ghorbani2019data}, etc. 

However, the aforementioned methods only provide a post-hoc solution and to truly provide interpretability, a more \textit{accesible} approach is required. Recently, there have been multiple concept-based models incorporate concepts during model training \cite{kim2018interpretability,zhou2018interpretable}. It is believed that explaining model predictions using abstract human-understandable ``concepts'' better \textbf{aligns} model's internal working with \textbf{human thought process}. Concepts can be thought of as abstract entities - shared across multiple samples providing a general model understanding. The general approach to train such models is to first map inputs to a concept space. Subsequently, alignment with the concepts is performed in the concept space and a separate model is learned on the concept space to perform the downstream task.

The ideal method to extract concepts from a dataset would be to manually curate and define what concepts best align with the requirements of stakeholders/end-users using extensive domain knowledge. This approach requires manual annotation of datasets and forces models to extract and encode only the pre-defined concepts as Concept Bottleneck Models \cite{koh2020concept,zaeem2021cause} do. However, with increasing dataset sizes, it becomes difficult to manually annotate each data sample, thus limiting the efficiency and practicality of such approaches \cite{yuksekgonul2022post}. 

As a result, many approaches incorporate unsupervised \textbf{concept discovery} for concept-based prediction models. One such architecture is Self Explaining Neural Networks (SENN) proposed in \cite{alvarez2018towards}. The concepts are extracted using a bottleneck architecture, and appropriate relevance scores to weigh each concept are computed in tandem using a standard feedforward network. The concepts and relevance scores are then combined using a network to perform downstream tasks (e.g. classification). Even though such concept-based explanations provide a clear explanation to understand neural machine intelligence, concept-based approaches are not without their faults. One critical problem we observed is that concepts learned across multiple domains using concept-based models are not consistent among samples from the same class, implying low \textbf{concept fidelity}. In addition, concepts are unable to generalize to new domains implying a lack of \textbf{concept-interoperability}.

In this paper, we propose a concept-learning framework with a focus on generalizable concept learning which improves concept interoperability across domains while maintaining high concept fidelity. Firstly, we propose a salient concept selection network that enforces representative concept extraction. Secondly, our framework utilizes self-supervised contrastive learning to learn domain invariant concepts for better interoperability. Lastly, we utilize prototype-based concept grounding regularization to minimize concept shifts across domains. Our novel methodology not only improves concept fidelity but also achieves superior concept interoperability, demonstrated through improved domain adaptation performance compared to SOTA self-explainable concept learning approaches. Our contributions are - (1)  We analyze the current SOTA self-explainable approaches for concept interoperability and concept fidelity when trained across domains - problems that have not been studied in detail by recent works. (2) We propose a novel framework that utilizes a \textit{salient concept selection network} to extract representative concepts and a self-supervised contrastive learning paradigm for enforcing domain-invariance among learned concepts. (3) We propose a prototype-based concept grounding regularizer to mitigate the problem of concept shift across domains. (4) Our evaluation methodology is the first to quantitatively evaluate the domain adaptation performance of self-explainable architectures and comprehensively compare existing SOTA self-explainable approaches.

%% file: 2_related.tex
\section{Related Work}
\label{sec:related}
\textbf{Related work on concept-level explanations}. Recent research has focused on designing concept-based deep learning methods to interpret how deep learning models can use high-level human-understandable concepts in arriving at decisions \cite{ghorbani2019towards,chen2019explaining,wu2020towards,koh2020concept,yeh2019completeness,mincu2021concept,huang2022conceptexplainer,leemann2022coherence,sinha2021perturbing,sinha2023understanding}. Such concept-based deep learning models aim to incorporate high-level concepts into the learning procedure. Concept priors have been utilized to align model concepts with human-understandable concepts \cite{zhou2018interpretable,murty2020expbert,chen2019explaining} and bottleneck models were generalized wherein any prediction model architecture can be transformed  \cite{koh2020concept,zaeem2021cause} by integrating an intermediate layer to represent a human-understandable concept representation. Similar work on utilizing CBMs for various downstream tasks include \cite{sawada2022concept,jeyakumar2021automatic,pittino2021hierarchical,bahadori2020debiasing}.

\textbf{Related work on self-supervised learning with images}.
Self-supervised learning \cite{xu2019self,saito2020universal} via pretext tasks has been demonstrated to learn high-quality domain invariant representations from images using a variety of transformations such as rotations \cite{xu2019self,gidaris2018unsupervised}. Self-supervised learning in image space fall into two major paradigms. The first approach generates multiple `views' or small transformations of the same image which preserve the inherent semantics. The transformations are usually small enough to not cause a significant shift in the intended and actual features in the latent space and are trained using a form of contrastive loss \cite{wang2021understanding}. The second paradigm attempts to view self-supervised feature learning as a puzzle-solving problem \cite{xu2019self}.

\textbf{Related work on automatic interpretable concept learning}.
Supervised concept learning requires the concepts of each training sample to be manually annotated, which is impossible with a moderately large dataset and the concepts are restricted to what humans can conceptualize. To alleviate such bottlenecks, automatic concept learning is becoming increasingly appealing. One dominant architecture is Self Explaining Neural Networks (SENN) proposed in \cite{alvarez2018towards}.
Several other popular methods have been proposed which automatically learn concepts are detailed \cite{kim2018interpretability,ghorbani2019towards,yeh2019completeness,wu2020towards,goyal2019explaining}.

\noindent \textbf{Comparision with existing work.}
Our work aims to address a challenge existing approaches face, concepts learned by self-explaining models may not be able to generalize well across domains, as the learned concepts are mixed with domain-dependent noise and less robust to light transformations due to a lack of supervision and regularization. Our proposed approach tackles this largely unsolved problem by designing a novel representative concept extraction framework and regularizes it using self-supervised contrastive concept learning and prototype-based grounding.

Concurrent to our work, BotCL \cite{wang2023learning} also proposes to utilize self-supervised learning to learn interpretable concepts. However, our approach is significantly different in both training and evaluation. We utilize multiple SOTA \textit{transformations} to learn distinct concepts, while BotCL only uses a very crude regularization by maximizing the similarity between samples from the same class during concept learning. Our evaluation framework is significantly more extensive and comprises of concept interoperability by evaluating performance across domains, while BotCL only uses task accuracy. Another work related to ours \cite{sawada2022concept} proposes to incorporate multiple unsupervised concepts in the bottleneck layer of CBMs in addition to supervised concepts which differs from our approach as we learn all concepts in a self-supervised manner, without supervision. Another concurrent work \cite{sawada2022c} attempts to utilize a modified autoencoder setup with a discriminator instead of a decoder and weak supervision using an object-detecting network (Faster RCNN) which is very specific to the autonomous driving datasets and is not generalizable.


%% file: 3_method.tex
\section{Methodology}
\label{sec:method}

In this section, we first provide a detailed description of our proposed learning pipeline, including (a) the Representative Concept Extraction (RCE) framework which incorporates a novel Salient Concept Selection Network in addition to the Concept and Relevance Networks, (b) Self-Supervised Contrastive Concept Learning (CCL) which enforces domain invariance among learned concepts, and (c) a Prototype-based Concept Grounding (PCG) regularizer that mitigates the problem of concept-shift among domains. We then provide details for the end-to-end training procedure with additional Concept Fidelity regularization which ensures concept consistency among similar samples.

\subsection{\underline{R}epresentative \underline{C}oncept \underline{E}xtraction}

\begin{figure}[h]
    \centering
    \includegraphics[width=0.85\linewidth]{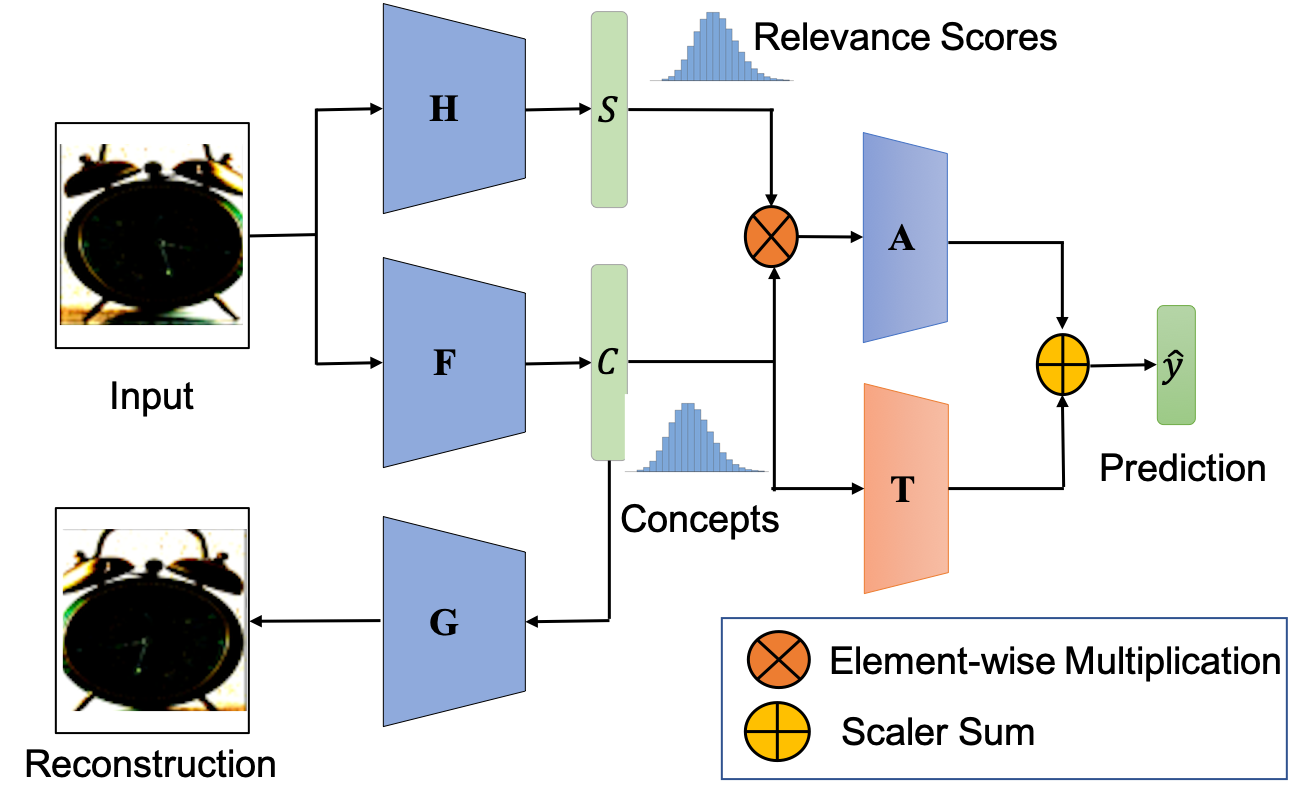}
    \caption{The proposed Representative Concept Extraction (RCE) framework. The networks $\mathbf{F}$ and $\mathbf{H}$ respectively extract concepts and associated relevance scores and $\mathbf{A}$ aggregates them. Network $\mathbf{G}$ reconstructs original input from the concepts while $\mathbf{T}$ selects the most representative concepts to the prediction.  }
    \label{fig:framework_rce}
\end{figure}

Figure \ref{fig:framework_rce} presents the proposed Representative Concept Extraction framework. For a given input sample $x \in \mathbb{R}^n$, the self-explainable concept learning framework learns a set of $K$ representative $d$-dimensional concepts $C = \{c_1,\cdots,c_K\} \in \mathbb{R}^d$ and relevance scores associated with the concepts $\mathcal{S} =\{s_1,\cdots,s_K\} \in \mathbb{R}^d$ for the downstream task. 

\noindent \textbf{Concept Network.}
The Concept Extraction Network consists of an encoder function $\mathbf{F}$, which maps from the input space to the concept representation space ($\mathbb{R}^n \rightarrow \mathbb{R}^d$). To preserve the maximum amount of information content in the concept representation, the entire network is modeled as an autoencoder with the decoder function $\mathbf{G}$ which maps from the concept representation space to the input space ($\mathbb{R}^d \rightarrow \mathbb{R}^n$). 

\noindent \textbf{Relevance Networks.} The Relevance Network function $\mathbf{H}$ is modeled similarly to the function $\mathbf{F}$, which maps from the input space to the concept representation space ($\mathbb{R}^n \rightarrow \mathbb{R}^d$). The relevance network outputs a score associated with each concept - encapsulating each concept's relevance to the prediction. Mathematically, the relevance network $\mathbf{H}$ ($\mathbb{R}^n \rightarrow \mathbb{R}^d$) outputs a set of score vectors $\mathcal{S} =\{s_1, .. s_k\}$ for an input sample $x$. 

\noindent \textbf{Salient Concept Selection Network.}
Note that approaches like \cite{alvarez2018towards} employ simple sparsity regularizations on the concept space to increase diversity and select representative concepts. However, we utilize a novel strategy that conditions the concept selection on the prediction performance. Effectively, utilizing a shallow network $\mathbf{T}$, which maps from the concept space to the prediction space ($\mathbb{R}^d \rightarrow \mathbb{R}$) \textit{selects} only those concepts that are \textit{most responsible} or \textit{salient} for prediction.

\noindent \textbf{Aggregation and Prediction.}
Subsequently, the concepts and the relevance scores are aggregated to perform the final prediction using the aggregation function $\mathbf{A}$ which maps from the concept space to the output prediction space ($\mathbb{R}^d \rightarrow \mathbb{R})$. Mathematically, the function $\mathbf{A}$ aggregates a given concept vector $\mathbf{F}(x)$ and relevance score vector $\mathbf{H}(x)$ respectively for a given sample $x$. A shallow fully connected network models the function $\mathbf{A}$. Note that the function $\mathbf{A}$ should be as shallow as possible to maximize interpretability. 

The final prediction is computed using a weighted sum of outputs from the Aggregation Network $\mathbf{A}$ and the Salient Concept Selection Network $\mathbf{T}$. Mathematically,
\begin{equation}
    \hat{y} = \omega_1* \mathbf{A}(\mathbf{F}(x)\odot\mathbf{H}(x)) + \omega_2*\mathbf{T}(\mathbf{F}(x)) 
\end{equation}
where $\odot$ is the element-wise product of the concept and relevance vectors. This weighted prediction strategy with tunable parameters $\omega_1$ and $\omega_2$ exerts greater control over concept selection. Note that higher values of $\omega_2$ enforce representative concept selection.

\noindent \textbf{Training Objective.}
As the Concept Network is modeled as an autoencoder, the training objective can be mathematically given by:
\begin{equation}
    \mathcal{L}_{rec} = L(x, \mathbf{G}(\mathbf{F}(x))) + \lambda |\mathbf{F}(x)|_1
    \label{eq:recons}
\end{equation}
Note that $\lambda$ is the strength of $L_1$ norm in Equation~\ref{eq:recons} -  that regularizes the concept space and prevents degenerate concept learning (such as all concepts being a unit vector). The reconstruction loss $\mathcal{L}_{rec}$ is composed of $L$ which quantifies the difference between an input sample $x$ and its reconstruction $\mathbf{G}(\mathbf{F}(x))$.

Note that as the network $\mathbf{F}$ is responsible for extracting representative concepts and the $\mathbf{H}$ is responsible for calculating the relevance of the concepts extracted by $\mathbf{F}$, they must be modeled by networks with similar complexity to avoid overfitting and learning of degenerate concepts.

The complete training objective of the Concept Extraction Framework where $\mathcal{L}$ is any prediction loss (such as Cross Entropy) is as follows:
\begin{equation}
    \mathcal{L}_{CE} = \mathcal{L}_{rec} + \mathcal{L}(y,\hat{y})  
\end{equation}

\subsection{Self-supervised \underline{C}ontrastive \underline{C}oncept \underline{L}earning}

\begin{figure}[h]
    \centering
    \vskip -8pt
    \includegraphics[width=0.85\linewidth]{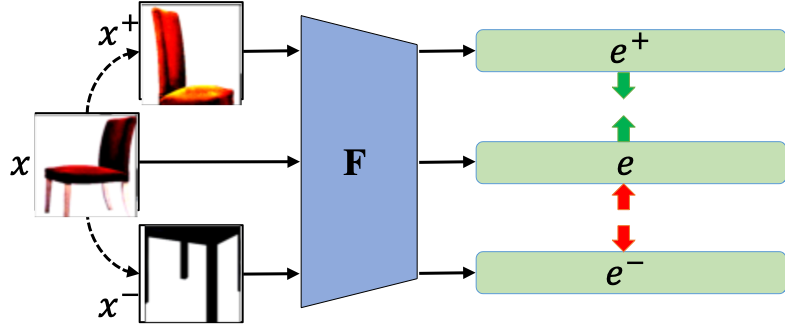}
    \vskip -8pt
    \caption{Self-supervised contrastive concept learning. Images sampled from a set of positive $X^+$ and negative samples $X^-$ associated with an anchor image $x$. Green arrows depict direction of maximizing similarity, red arrows depict direction of minimizing similarity.}
    \vskip -10pt
    \label{fig:framework_2}
\end{figure}

Even though the RCE framework generates representative concepts, the concepts extracted are adulterated with \textit{domain noise} thus limiting their generalization. In addition, with limited training data, the concept extraction process is not robust. Self-supervised learning contrastive training objectives are the most commonly used paradigm \cite{thota2021contrastive} for learning robust visual features in images. We incorporate self-supervised contrastive learning to learn domain invariant concepts, termed CCL. 

\noindent \textbf{Contrastive Sampling Procedure.} The underlying idea revolves around utilizing multiple strong transformations of an input sample $x_i$ (anchor) and maximizing the similarity between their representations and minimizing the similarity between non-related transformations in the concept space, as shown in Figure \ref{fig:framework_2}. Mathematically, given an image sample $x_i$ and the Concept Network $\mathbf{F}$, a set of transformations $T = \{t_1, t_2, ... t_n \}$, Contrastive learning begins by imputing a set of positive samples wrt $x_i$ $X^{+} = \{t_1(x_i), t_2(x_i) ... t_n(x_i) \}$ and negative samples wrt $x_j$ $X^{-} = \{t_1(x_j), t_2(x_j) ... t_n(x_j) \}$. Note that the negative samples are not sampled from transformations of $x_i$ but another sample $x_j$ such that $i \neq j$. The concept representations wrt the positive and negative sets given a Concept Network $\mathbf{F}$ are  $E^{+} = \{e^+_i = \mathbf{F}(x_i)~~\forall x_i\in X^{+}\}$ and $E^{-} = \{e^-_i = \mathbf{F}(x_i)~~\forall x_i\in X^{-}\}$ respectively.

\noindent \textbf{Self-Supervised Training Objective.} The extent of similarity is adjusted using a tunable hyperparameter $\tau$ (\textit{temperature}), which controls the penalty on both positive and negative samples. Formally, the self-supervised loss $\mathcal{L}_{ssl}$ parameterized by an anchor image sample's concept representation $e = \mathbf{F}(x)$), its associated positive and negative sets' concept representations $E^+$ and $E^-$ can be formulated as Equation~\ref{eq:ssl}.
\begin{equation}
    \label{eq:ssl}
    \mathcal{L}_{ssl}= -\log (\frac{\text{exp}(s(e,e^+)/\tau)}{\sum^{|E^{-}|}\text{exp}(s(e,e^-)/{\tau}))}) 
\end{equation}
where $e^+ \in E^+,e^- \in E^-$ and $s$ is any similarity function.

\subsection{\underline{P}rototype-based \underline{C}oncept \underline{G}rounding}

\begin{figure}[t]
    \centering
    \includegraphics[width=0.8\linewidth]{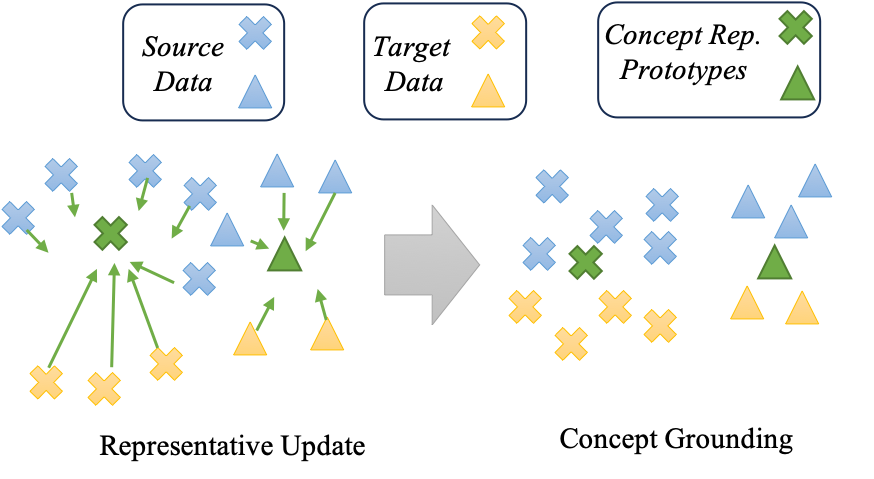}
     \vskip -10pt
    \caption{Prototype-based concept grounding (PCG). Concept grounding ensures the concept representations learned from both source and target domains are \textit{grounded} to a representative concept representation prototype (Green).}
    \vskip -11pt
    \label{fig:framework_3}
\end{figure}
Ideally, concepts should be invariant entities shared among samples from similar classes and aligned across domains. However, due to imbalanced data and domain noise, models without explicit regularization learn significantly divergent concept representations for similar samples across domains, a phenomenon termed \textit{concept-shift}. For proper concept alignment, it is important to ensure concept representations associated with samples of the same class from different domains are as close as possible. To achieve this, we utilize a prototype as an anchor, which \textit{grounds} concept representations from multiple domains and reduces \textit{concept-shift} during training. An illustration of concept grounding is presented in Figure \ref{fig:framework_3}. The blue and yellow data points correspond to concept representations for a class in the source and target domains respectively while the crosses and triangles represent different types of concepts. Our objective is to ground the source and target concept representations using a \textit{concept representation prototype} (shown in green). Note that the training data $X$ in our setting is a set of abundant samples from a source domain, $X^s$, and non-abundant samples from a target domain, $X^t$, i.e., $X = X^s \cup X^t$. Our prototype-based concept grounding method (PCG) utilizes a dynamically updated bank of concept representation prototypes to enforce concept alignment during training. 
The \textit{concept bank} is constructed with concept representations of randomly sampled data points for each class from both source and target domains. Let $N$ be the set of classes in the task. We sample a set of samples $S^s \subset X^s$ such that $S^s = \cup_{c=1}^N\mathcal{S}_c^s$ where $S^s_c$ is a set of randomly selected samples belonging to class $c$ from the source domain. Similarly, set $S^t\subset X^t$ is sampled from the target domain such that $S^t = \cup_{c=1}^N\mathcal{S}_c^t$. The representative concept prototype corresponding to a class $c$, $\mathcal{C}_c$, is updated after every training step with a weighted sum of the source and target concept prototypes associated with $S^t$ and $S^s$:
\begin{align}
    \label{eq:rep-cc}
    \begin{split}
      \mathcal{C}_c \leftarrow \frac{\mu}{|S^s_c|}\sum_{x \in S^s_c}~\mathbf{F}(x) + \frac{(1-\mu)}{|S^t_c|}\sum_{x \in S^t_c}~\mathbf{F}(x)
    \end{split}
\end{align}
where $\mu$ is a tunable hyperparameter used to control the extent of concept shift. Note that the higher the $\mu$, the more concepts will be grounded to the source domain. The grounding concept code bank $\mathcal{C} = \{\mathcal{C}_c, ~\forall~c\in~N\}$ is used to supervise the concept representation learning as follows:
\begin{equation}
    \mathcal{L}_{grnd} = L(\mathbf{F(x)}, \mathcal{C})
\end{equation}
where \(L\) is the same loss function as shown in Formula \ref{eq:recons}, which can be implemented as Mean Square Error. 

\noindent \textbf{Concept Fidelity Regularization.}
Concept fidelity attempts to enforce the similarity of concepts through a similarity measure $s (\cdot, \cdot)$ of data instances from the same class in the same domain. Formally,
\begin{equation}
\label{eq:parity}
    \mathcal{L}_{fid} = s(\mathbf{F}(x_i),\mathbf{F}(x_j))~~~\text{for}~~ y_i = y_j
\end{equation}

\subsection{End-to-end Composite Training}
Overall, the training objective can be formalized as a weighted sum of  CCL and PCG objectives:
\begin{equation}
    \mathcal{L}_{CL} = \mathcal{L}_{ssl} + \lambda_1*\mathcal{L}_{grnd} + \lambda_2*\mathcal{L}_{fid}
\end{equation}
where $\lambda_1$ and $\lambda_2$ are tunable hyperparameters controlling the strength of contrastive learning and prototype grounding regularization.
The end-to-end training objective can be represented as:
\begin{equation}
    \mathcal{L}_{CE} + \beta* \mathcal{L}_{CL}
\end{equation}
The tunable hyperparameter $\beta$ controls the effect of generalization and robustness on the RCE framework. Note that a higher value of $\beta$ makes the concept learning procedure brittle and unable to adapt to target domains. However, a very low value of $\beta$ makes the concept learning procedure overfit on the source domain, implying a tradeoff between concept generalization and performance.

%% file: 4_experiments.tex
\section{Experiments}
\label{sec:experiments}

\begin{table}[h]
\centering
\resizebox{0.45\textwidth}{!}{
\begin{tabular}{ccccc}
\hline
Method   & Explainable & Prototypes & Interoperability & Fidelity  \\
\hline

S+T      &      \xmark  &    \xmark     &  \xmark     &   \xmark    \\
SENN      &     \redcheck    &  \redcheck   &    \xmark   & \xmark      \\

DiSENN     &     \redcheck    &   \redcheck    &   \redcheck     &  \xmark     \\

BotCL      &    \redcheck    &   \redcheck   &  \xmark  &   \redcheck    \\

UnsupCBM   &   \redcheck     & \xmark    & \xmark    &  \xmark     \\

\textbf{Ours}       &    \redcheck   & \redcheck   &  \redcheck  & \redcheck      \\
\hline                                                
\end{tabular}}
\vskip -5pt
\caption{A summary of salient features of our method as compared to the baselines considered. The column `Explainable' shows whether the method is inherently explainable without any post-hoc methodologies. Column `Prototypes' depicts if a method can explain predictions by selecting prototypes from the train set, `Interoperability' shows if learned concepts maintain consistency across domains and `Fidelity' depicts if the method maintains intra-class consistency among learned concepts.}
\vskip -15pt
\label{tab:baseline-summary}
\end{table}

\begin{table*}[t]
    \centering
    \resizebox{0.85\textwidth}{!}{\begin{tabular}{ccccccccccccc}
        \hline
        & A $\rightarrow$ C & A $\rightarrow$ P & A $\rightarrow$ R &  C $\rightarrow$ A & C $\rightarrow$ P & C $\rightarrow$ R & P $\rightarrow$ A & P $\rightarrow$ C & P $\rightarrow$ R & R $\rightarrow$ A  & R $\rightarrow$ C & R $\rightarrow$ P \\
        \hline
          S+T &  54.0 & 73.1 & 74.2 & 57.6 & 72.3 & 68.3 & 63.5 & 53.8 & 73.1 & 67.8 & 55.7 & 80.8 \\
         \hline
          SENN &  52.5 & 73.1 & 74.2 & 57.6 & 72.3 & 68.3 & 59.5 & 53.8 & 73.1 & 66.3 & 55.7 & 80.8 \\
          DiSENN &  48.5 & 69.2 & 70.1 & 52.5 & 69.1 & 66.1 & 58.8 & 51.2 & 70.3 & 64.9 & 52.3 & 77.0 \\
         BotCL &  53.1 & 72.8 & 74.0 & \textbf{58.2} & 70.4 & 67.9 & 58.4 & 52.1 & 72.6 &  65.3 & 56.3 & 78.2  \\
         UnsupCBM &  54.0 & 73.1 & 74.2 & 57.6 & 72.3 & 68.3 & 63.5 & 53.8 & 73.1 & \textbf{67.8} & 55.7 & 80.8 \\
         \hline
         RCE &  52.5 & 73.1 & 74.2 & 57.6 & 72.3 & 68.3 & \textbf{63.5} & 53.8 & 73.1 & 67.8 & 55.7 & 80.8 \\
         RCE+PCG & 55.2 & 73.1 & 74.0 & 57.9 & 71.2 & 68.1 & 58.1 & 53.6 & 73.2 & 66.9 & 56.1 & 80.3  \\  
         \textbf{RCE+PCG+CCL} & \textbf{58.7} &\textbf{73.7} & \textbf{75.0} & 58.0 & \textbf{71.9} & \textbf{68.9} & 62.1 & \textbf{55.4} & \textbf{74.8}& 67.2 & \textbf{60.2} & \textbf{81.3}   \\  
         \hline
    \end{tabular}}
    \vskip -8pt
    \caption{Domain generalization performance for the Office-Home Dataset with domains Art (A), Clipart (C), Product (P) and Real (R). } 
    \vskip -10pt
    \label{tab:officehome-acc}
\end{table*}

\begin{table*}[t]
    \centering
    \resizebox{0.7\textwidth}{!}{\begin{tabular}{cccccccc||cc}
        \hline
        \multicolumn{8}{c||}{DomainNet} & \multicolumn{2}{c}{VisDA} \\ 
        \hline
        & R $\rightarrow$ C & R $\rightarrow$ P & P $\rightarrow$ C & C $\rightarrow$ S & S $\rightarrow$ P & R $\rightarrow$ S & P $\rightarrow$ R &  R $\rightarrow$ 3D & 3D $\rightarrow$ R \\
        \hline
          S+T & 60.0  & 62.2 & 59.4 & 55.0  & 59.5 & 50.1 & 73.9 & 79.8 &  49.4 \\
         \hline
          SENN & 59.2  & 60.1 & 57.2 & 53.8  & 56.1 & 49.0 & 72.4 & 79.6 & 49.2\\
          DiSENN & 57.3  & 58.1 & 55.3 & 51.2  & 55.1 & 47.4 & 71.0 & 78.1 &48.1\\
         BotCL & 60.0 & 60.1 & 57.2 & 53.8  & 56.1 & 49.0 & 72.4 & 80.2 & 49.8\\
         UnsupCBM & 60.0  & 62.2 & 59.4 & 55.0  & \textbf{59.5} & 50.1 & 73.9 & 80.3& 49.9\\
         \hline
         RCE & 59.2  & 60.1 & 57.2 & 53.8  & 56.1 & 49.0 & 72.4 & 79.6 & 49.2\\
         RCE+PCG & 60.8  & 59.9  & 59.9 &  54.6 & 58.9 & 51.6 &  73.6 & 81.3 & 49.5\\  
         \textbf{RCE+PCG+CCL} & \textbf{61.2}  & \textbf{60.5} &  \textbf{62.9} & \textbf{55.0} & 59.1 & \textbf{52.1} & \textbf{74.2}  & \textbf{82.4} & \textbf{53.4} \\  
         \hline
    \end{tabular}}
    \vskip -8pt
    \caption{Domain generalization performance for the [Left] DomainNet dataset with domains Real (R), Clipart (C), Picture (P), and Sketch (S) and [Right] VisDA dataset with domains Real (R) and 3-Dimensional visualizations (3D).  } 
    \vskip -15pt
    \label{tab:domainnet-acc}
    
\end{table*}

\begin{table}[t]
    \centering
    \resizebox{0.49\textwidth}{!}{\begin{tabular}{c|c|c|c|c|c|c}
    \hline
          & M $\rightarrow$ U & M $\rightarrow$ S & U $\rightarrow$ M & U $\rightarrow$ S & S $\rightarrow$ M & S $\rightarrow$ U  \\
         \hline
         S+T & 0.54 & 0.16 & 0.74 & 0.13 & 0.92 & 0.65 \\
         SENN & 0.43 & 0.11 & 0.73 & 0.09 & 0.92 & 0.64 \\
         DiSENN   & 0.43 & 0.11 & 0.73 & 0.09 & 0.92 & 0.64  \\
         BotCL & 0.58 & 0.14 & 0.17 & 0.12 & 0.38 & 0.51 \\
         UnsupCBM  & 0.54 & 0.16 & 0.74 & 0.13 & 0.92 & 0.65 \\
         \hline
         RCE & 0.43 & 0.11 & 0.73 & 0.09 & 0.92 & 0.64 \\
         RCE+PCG & 0.58 & 0.23 & 0.79 & 0.19 & 0.94 & 0.71  \\
         \textbf{RCE+PCG+CCL} & \textbf{0.60} & \textbf{0.23} & \textbf{0.81} & \textbf{0.20} & \textbf{0.95} & \textbf{0.71}  \\
         \hline
    \end{tabular}}
    \vskip -6pt
    \caption{Domain generalization performance for the Digit datasets with domains MNIST (M), USPS (U) and SVHN (S).  In addition, we also report the results of multiple source domain adaptation to the target domains in the Appendix.}
    \vskip -15pt
    \label{tab:digit-acc}
\end{table}

\subsection{Datasets and Networks}
We consider four widely used task settings commonly utilized for domain adaptation. The task in each of the following settings is classification. 
\begin{itemize}[leftmargin=*, parsep=0pt, itemsep=0pt, topsep=0pt]
    \item \textbf{Digits}: This setting utilizes MNIST and USPS \cite{lecun1998gradient,hull1994database} with Hand-written images of digits and Street View House Number Dataset (SVHN) \cite{netzer2011reading} with cropped house number photos.
    \item \textbf{VisDA-2017} \cite{peng2017visda}: contains 12 classes of vehicles sampled from Real (R) and 3D domains.


    \item \textbf{DomainNet} \cite{venkateswara2017deep}: contains 126 classes of objects (clocks, bags, etc.) sampled from 4 domains - Real (R), Clipart (C), Painting (P), and Sketch (S).
    
    \item \textbf{Office-Home} \cite{peng2019moment}: Office-Home contains 65 classes of office objects like calculators, staplers, etc. sampled from 4 different domains - Art (A), Clipart (C), Product (P), and Real (R). 
\end{itemize}

\noindent \textbf{Network Choice:} 
For Digits, we utilize a modified version of LeNet \cite{lecun1998gradient} which consists of 3 convolutional layers for digit classification with ReLU activation functions and a dropout probability of 0.1 during training. 
For all other datasets we utilize a ResNet34 architecture similar to \cite{yu2023semi} and initialize it with pre-trained weights from Imagenet1k. For details, refer Appendix. 

\begin{figure}[h]
    \centering
    \includegraphics[width=0.3\textwidth]{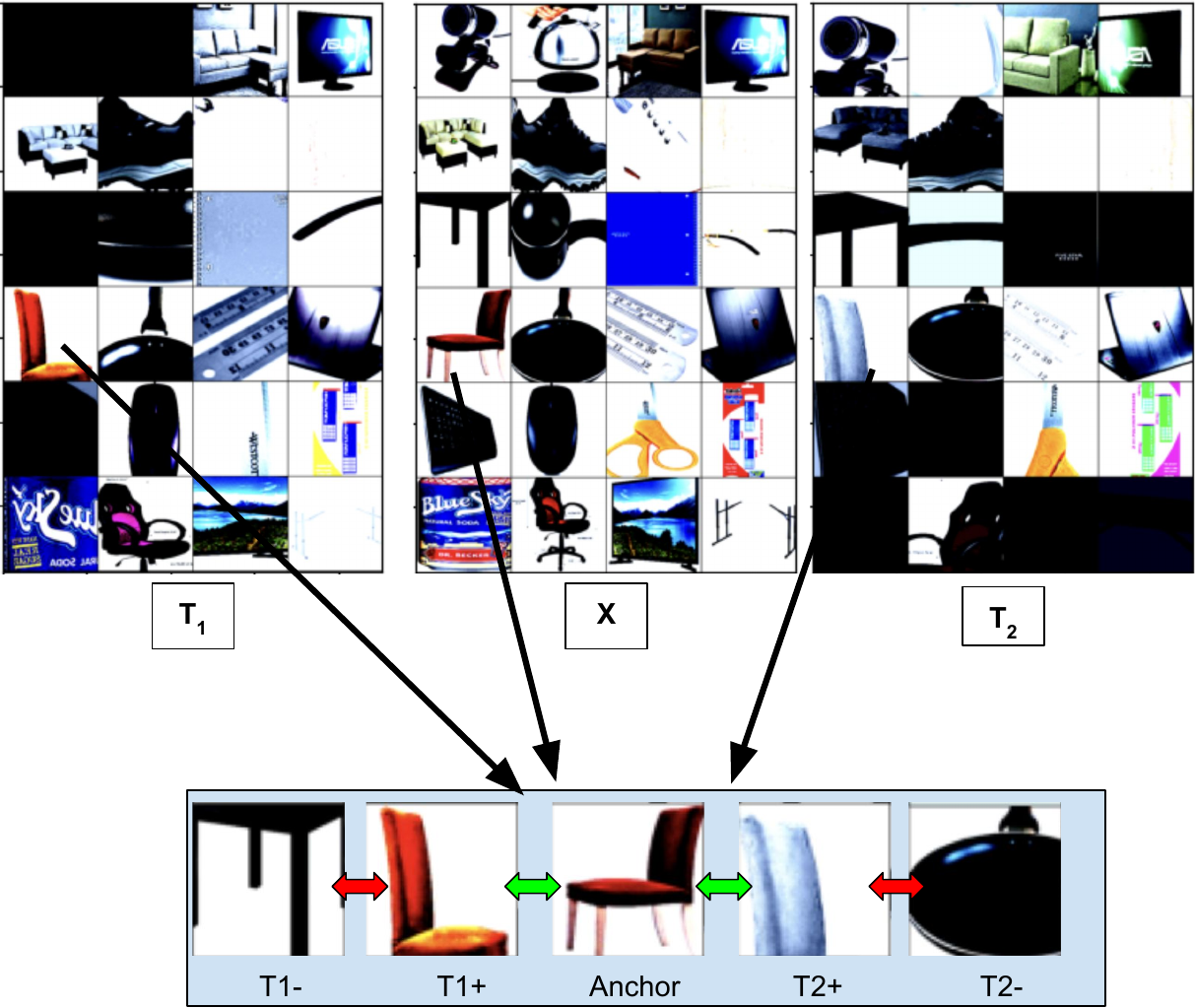}
    \caption{Schematic overview of proposed SimCLR transformations for OfficeHome dataset from the Product(P) domain. Note that green arrows depict maximizing similarity while red arrows depict minimizing similarity in concept space. Transformation sets $T_1+$ and $T_2+$ comprise images transformed from chair while $T_1-$ and $T_2-$ consist of images transformed from non-chair classes.}
    \vskip -12pt
    \label{fig:simclr-eg}
\end{figure}

\noindent\textbf{Baselines.}
We start by comparing against standard non-explainable NN architectures - the S+T setting as described in \cite{yu2023semi}. Next, we compare our proposed method against 5 different self-explaining approaches. As none of the approaches specifically evaluate concept generalization in the form of domain adaptation, we replicate all approaches. \textbf{SENN} and \textbf{DiSENN} utilize a robustness loss calculated on the Jacobians of the relevance networks with DiSENN utilizing a VAE as the concept extractor. \textbf{BotCL} \cite{wang2023learning} also proposes to utilize contrastive loss but uses it for position grounding. Similar to BotCL, Ante-hoc concept learning \cite{sarkar2022framework} uses contrastive loss on datasets with known concepts, hence we do not explicitly compare against it. Lastly, \textbf{UnsupervisedCBM} \cite{sawada2022concept} uses a mixture of known and unknown concepts and requires a small set of known concepts. For our purpose, we provide the one-hot class labels as known concepts in addition to unknown. A visual summary of the salient features of each baseline is depicted in Table~\ref{tab:baseline-summary}.

\subsection{Hyperparameter Settings}
\label{sec:hp-setting}
\noindent \textbf{RCE Framework}: We utilize the Mean Square Error as the reconstruction loss and set sparsity regularizer $\lambda$ to 1e-5 for all datasets. The weights $\omega_1$ = $\omega_2$ = 0.5 are utilized for digit, while they are set at  $\omega_1$ = 0.8 and $\omega_2$ = 0.2 for object tasks.

\noindent \textbf{Learning}: We utilize the \textit{lightly}\footnote{https://github.com/lightly-ai/lightly} library for implementing SimCLR transformations \cite{chen2020simple}.  We set the temperature parameter ($\tau$) to 0.5 by default \cite{xu2019self} for all datasets. The hyperparameters for each transformation are defaults utilized from SimCLR. The training objective is Contrastive Cross Entropy (NTXent) \cite{chen2020simple}. Figure~\ref{fig:simclr-eg} depicts an example of various transformations along with the adjudged positive and negative transformations.
For the training procedure, we utilize the SGD optimizer with momentum set to 0.9 and a cosine decay scheduler with an initial learning rate set to $0.01$. We train each dataset for 10000 iterations with early stopping. The regularization parameters of $\lambda_1$ and $\lambda_2$ are set to 0.1 respectively. For Digits, $\beta$ is set to 1 while it is set to 0.5 for objects.  For further details, refer to Appendix.

\vskip -10pt
\subsection{Evaluation Metrics}
\vskip -3pt
We consider the following evaluation metrics to evaluate each component of the concept discovery framework. 

\begin{itemize}[leftmargin=*, parsep=0pt, itemsep=0pt, topsep=0pt]
    \item \textbf{Generalization}: We start by quantitatively evaluating the quality of concepts learned by measuring how well the learned concepts can generalize to new domains. To achieve this, we compare our proposed method against the aforementioned baselines on domain adaptation settings. 
    
    \item \textbf{Concept Fidelity}: To evaluate consistency in the learned concepts, we compute the intersection over union of the concept sets associated with for two data points $x_i$ and $x_j$ from same class as defined in Equation~\ref{eq:jsi}:
\begin{equation}
     \text{Fidelity score} = {|C^{x_i} \cap C^{x_j}|}~\text{\slash}~{|C^{x_i} \cup C^{x_j}|}
     \label{eq:jsi}
\end{equation}
    
\end{itemize}

\subsection{Genenralization Results}
Tables~\ref{tab:officehome-acc},~\ref{tab:domainnet-acc}, and~\ref{tab:digit-acc} report the domain adaptation results on the OfficeHome, DomainNet, VisDA and the Digit datasets, respectively. The notation X$\rightarrow$Y represents models trained on X as the source domain (with abundant data) and Y as the target domain (with limited data) and evaluated on the test set of domain $Y$. The best statistically significant accuracy is reported in bold. 
The last three rows in all the tables list the performance of the \textbf{RCE} framework, RCE trained with regularization (\textbf{RCE+PCG}), and RCE trained with both regularization and contrastive learning paradigm (\textbf{RCE+PCG+CCL}). 

\noindent \textbf{Comparision with baselines.} The first row in each table lists the performance of a standard Neural Network trained using the setting described in \cite{yu2023semi} (S+T). As a standard NN is not inherently explainable, we consider this setting as a baseline to understand the upper bound of the performance-explainability tradeoff. 

The second and third rows in each table lists the performance of SENN and DiSENN respectively. SENN performs worse than S+T setting in almost all settings, except in a handful of settings where the performance matches S+T. This is expected, as SENN is formulated as an overparameterized version of a standard NN with regularization. Recall that DiSENN replaces the autoencoder in SENN with a VAE, and as such is not generalizable to bigger datasets without domain engineering. DiSENN performs the worst among all approaches for all datasets due to poor VAE generalization. 

Recall that UnsupervisedCBM is an improved version of SENN architecture with a discriminator in addition to the aggregation function. In most cases, it performs slightly better than SENN and is at par with S+T. However, in particular cases in OfficeHome data (R$\rightarrow$A) and DomainNet (S$\rightarrow$P), UnsupCBM performs the best. We attribute this result to two factors: first, the Art (A) and Sketch (S) domains are significantly different from Real (R) and Picture (P) domains due to both of the former being hand-drawn while the latter being photographed as mentioned in \cite{yu2023semi}. Second, the use of a discriminator as proposed in UnsupervisedCBM helps enforce domain invariance in those particular cases.

BotCL explicitly attempts to improve concept fidelity and applies contrastive learning to \textit{discover} concepts. However, the contrastive loss formulation is rather basic and they never focuses on domain invariance. BotCL's performance is similar to S+T for the most part except in OfficeHome data (C$\rightarrow$A), where it just outperforms all other approaches. One possible reason is that Clipart domain is significantly less noisy, and hence basic transformations in BotCL work well.

As the last row demonstrates, our proposed framework RCE+PCG+CCL outperforms all baselines on a vast majority of the settings across all four datasets and is comparable to SOTA baselines in the other settings. 

\noindent \textbf{Ablation studies.} 
We also report the performance corresponding to various components of our proposed approach. We observe that the performance of RCE is almost identical to SENN, which is expected as there is very weak regularization in both cases. In almost all cases, adding prototype-based grounding regularization (RCE+PCG) improves performance over RCE while models trained with both PCG regularization and contrastive learning (RCE+PCG+CCL) outperform all approaches on a vast majority of settings across all datasets. Note that the setting RCE+CCL is not reported, as it defeats the fundamental motivation of maintaining concept fidelity.

\noindent \textbf{Effect of number of concepts and dimensions.} We observe that there are no significant differences in performance over varying number of concepts or dimensions. For all results reported, the number of concepts is set to number of classes in the dataset and their dimension is set to 1. For results on varying number of concepts and dimensions - refer Appendix. 

\vskip -5pt

\subsection{Concept Fidelity}
\vskip -3pt
As RCE framework is explicitly regularized with a concept fidelity regularizer and grounded using prototypes, we would expect high fidelity scores. Table~\ref{tab:jsi} lists the fidelity scores for the aforementioned baselines and our proposed method. Fidelity scores are averaged for each domain when taken as target (e.g. for domain (A) in DomainNet, the score is average of C$\rightarrow$A, P$\rightarrow$A and R$\rightarrow$A).
As expected, our method and BotCL, both with specific fidelity regularization outperform all other baseline approaches. Our method outperforms BotCL on most settings, except when the target domains are Art in DomainNet and Clipart in OfficHome due to siginificant domain dissonance.

\begin{table}[h]
    \centering
    \resizebox{1.02\columnwidth}{!}{%
    \begin{tabular}{c | c|c|c  | c|c|c|c |  c|c|c|c}
    \hline
     & \multicolumn{3}{c|}{\textbf{Digit}} & \multicolumn{4}{c|}{\textbf{DomainNet}} & \multicolumn{4}{c}{\textbf{OfficeHome}}\\
    \hline
         & M &  U &  S  &  A & C & P & R & C & P & R & S\\
         \hline
          SENN & 0.81 &  0.74 & 0.61   & 0.21 & 0.24 & 0.26 & 0.30 & 0.31 & 0.27 & 0.29 & 0.30  \\
         
          DSENN &  0.79 &  0.71 & 0.63    & 0.14 & 0,22 & 0.21 & 0.27 & 0.29 & 0.23 & 0.29 & 0.32  \\
          
          BotCL & 0.93 & \textbf{0.94}  & 0.89  & \textbf{0.49} & 0.55 & 0.51 & 0.58 & \textbf{0.73} & 0.66  & 0.61 & 0.64 \\
          
          UnsupCBM & 0.79 &  0.74 &  0.63   & 0.21 & 0.24 & 0.26 & 0.30   & 0.31 & 0.27 & 0.29 & 0.30  \\
          \hline
          
          RCE & 0.86 &  0.80 & 0.73   & 0.39 & 0.50 & 0.45 & 0.42 & 0.54 & 0.49 & 0.50 & 0.49 \\
          
          RCE+PCG &  0.94 &  0.94 & 0.89   & 0.47 & 0.56 & 0.51 & 0.59 & 0.70 & 0.67 & 0.61 & 0.63 \\
          
          \textbf{RCE+PCG+CCL} &  \textbf{0.94} &  0.94 & \textbf{0.89}    & 0.47 & \textbf{0.55} & \textbf{0.52} & \textbf{0.59} & 0.71 & \textbf{0.68} & \textbf{0.63} & \textbf{0.64} \\
          \hline
    \end{tabular}%
    }
    \vskip -6pt
    \caption{Average Intra-class Concept Fidelity scores for each domain for all settings where the domain is target. The columns show the domains in each dataset. For complete table, refer Appendix.}
    \vskip -15pt
    \label{tab:jsi} 
\end{table}

\subsection{Qualitative Visualization}
\begin{figure}[h]
    \centering
    \includegraphics[width=0.4\textwidth]{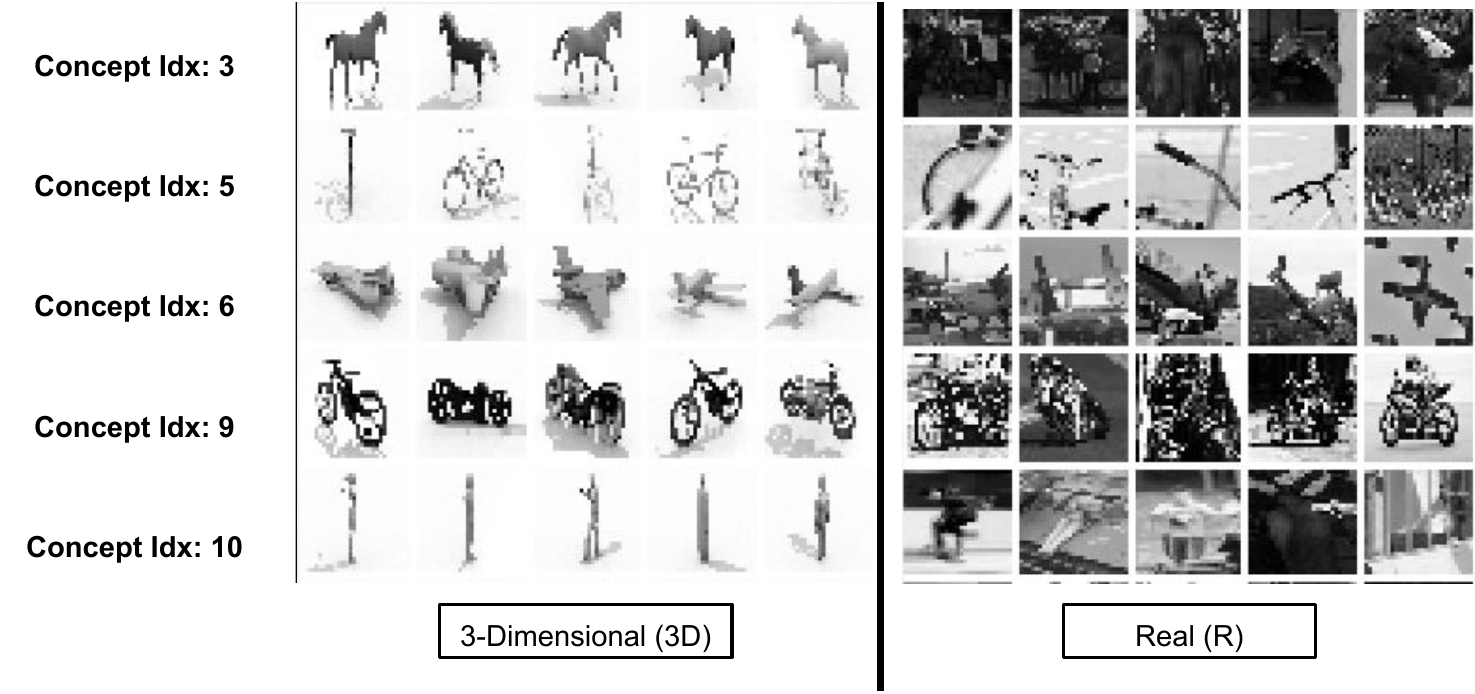}
    \includegraphics[width=0.4\textwidth]{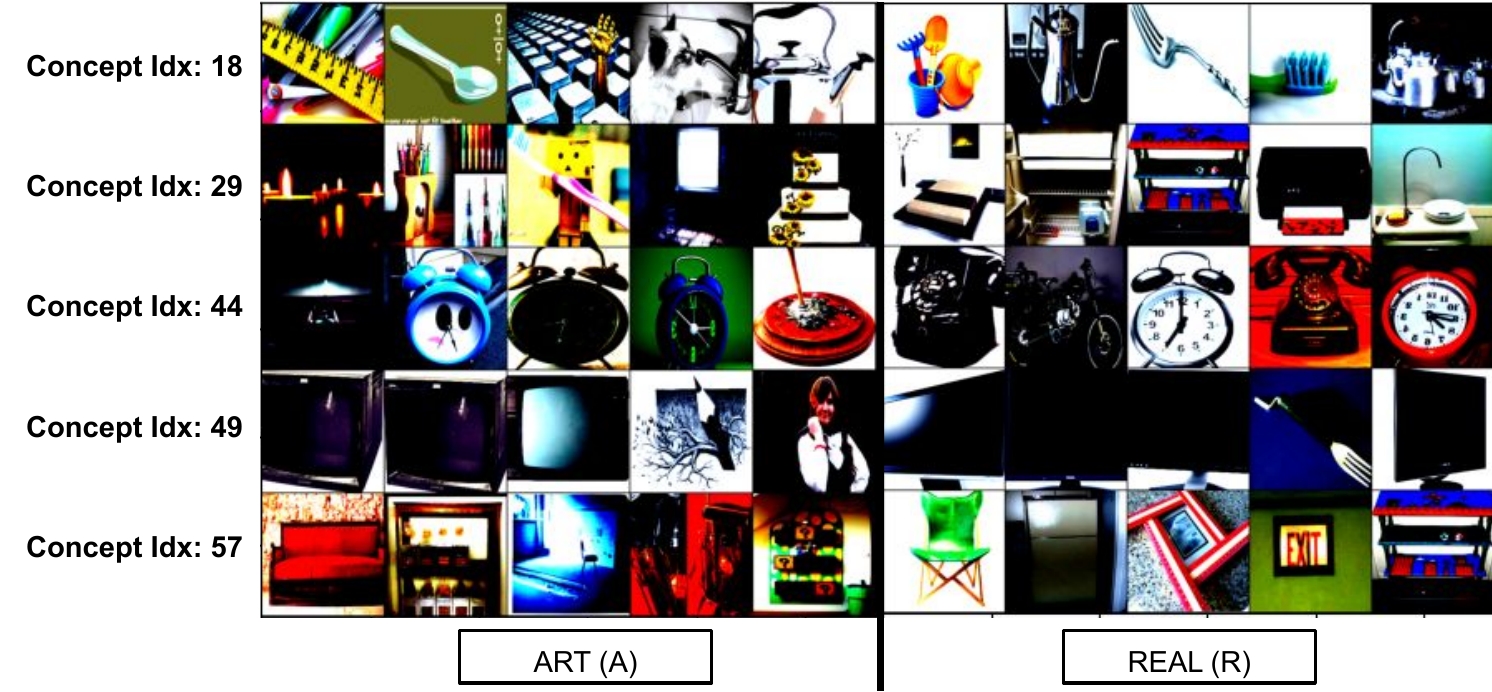}
    \caption{Top-5 most important prototypes associated with randomly chosen concepts on a model trained using our methodology on the VisDA [TOP] and OfficeHome [BOTTOM] dataset for the 3D $\rightarrow$ Real and  Art (A) $\rightarrow$ Real (R) domains respectively. The prototypes on the left are chosen from the training set of the source domain and the ones on the right are chosen from the target domain. As can be seen, in the VisDA dataset Concept \#6 captures samples with wings - namely airplanes and oddly shaped cars while in OfficeHome, Concept \#44 captures training samples with rounded faces in both domains - including alarm clocks, rotary telephones, etc. Similarly, Concept \#29 captures flat screens - TVs, and monitors.}
    \vskip -13pt
    \label{fig:domain-alignment}
\end{figure}

\noindent \textbf{Domain Alignment.}
We consider the extent to which the models trained using both concept grounding and contrastive learning maintain concept consistency not only within the source domain but also across the target domain as well. To understand what discriminative information is captured by a particular concept, Figure~\ref{fig:domain-alignment} shows the most important prototypes selected from the training set of both the source and target domains corresponding to five randomly selected concepts. We observe that prototypes explaining each concept are visually similar. For more results, refer Appendix.

\begin{figure}[h]
    \centering
    \includegraphics[width=0.45\textwidth]{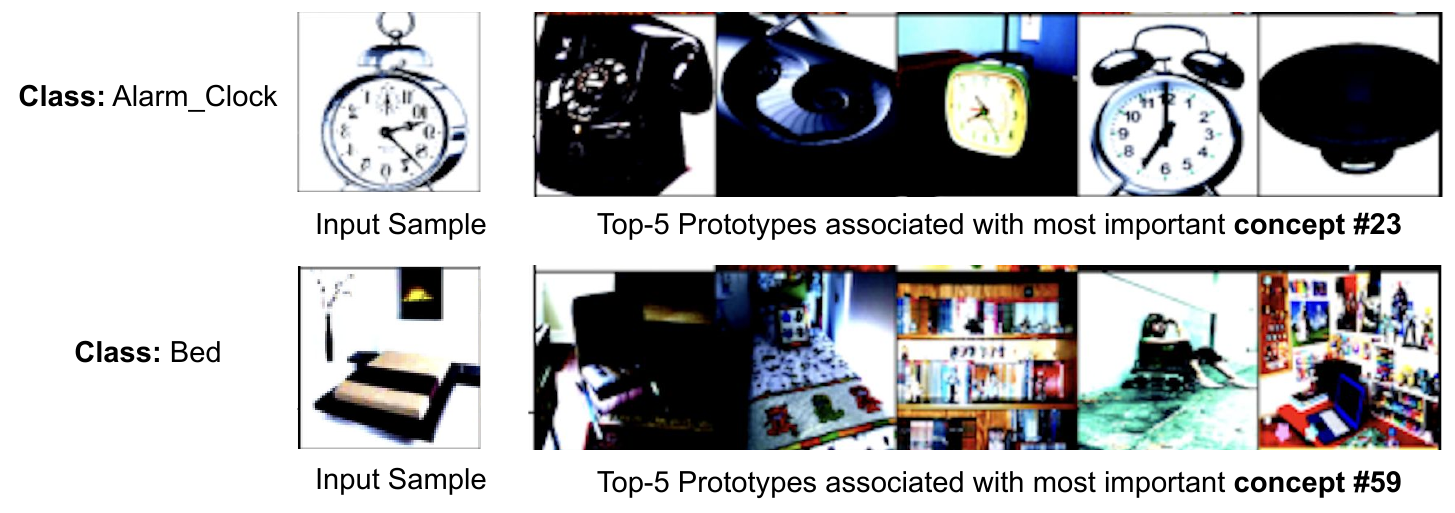}
    \caption{We demonstrate the top-5 most important prototypes associated with the highest activated concept for a particular correctly predicted input sample from target domain.  The top sample is correctly predicted - Alarm\_Clock - the prototypes associated to the most important concept are distinctly circular objects. Similarly, prototypes associated with sample from bed class are mostly flat.}
    \vskip -12pt
    \label{fig:class-prototypes}
\end{figure}

\noindent \textbf{Explanation using prototypes.}
For a given input sample, we also plot the prototypes associated to the highest activated concept, i.e., the important concept. Figure~\ref{fig:class-prototypes} shows the prototypes associated with the concepts most responsible for prediction (highest relevance scores). As can be seen, the prototypes possess distinct features, for eg., they capture round face of alarm clock. More results are reported in Appendix.

%% file: 6_conclusion.tex
\section{Conclusion}
\label{sec:conclusion}
 \vskip -3pt
In this paper, we discuss a fairly less-studied problem of \textit{concept interoperability} which involves learning domain invariant concepts that can be generalized to similar tasks across domains. Next, we introduce a novel Representative Concept Extraction framework that improves on present self-explaining neural architectures by incorporating a Salient Concept Selection Network. We propose a Self-Supervised Contrastive Learning-based training paradigm to learn domain invariant concepts and subsequently propose a Concept Prototype-based regularization to minimize concept shift and maintain high fidelity. Empirical results on domain adaptation performance and fidelity scores show the efficacy of our approach in learning generalizable concepts and improving concept interoperability. Additionally, qualitative analysis demonstrates that our methodology not only learns domain-aligned concepts but is also able to explain samples from both domains equally well. We hope our research helps the community utilize self-explainable models in domain alignment problems in the future.

%% file: appendix.tex
\newpage
\section{Appendix}

\noindent \textbf{Structure of Appendix}

\noindent Following discussions from the main text, the Appendix section is organized as follows:
\begin{itemize}[leftmargin=*]
    \itemsep0em
    \item Dataset descriptions and visual samples 
    \item Detailed discussion around RCE and algorithmic details for CCL and PCG (Pseudocode) 
    \item More experimental results on key hyperparameters utilized in RCE and PCG
    \item Concept Fidelity Analysis
    \item Details on Baseline Replication
    \item Additional visual results - selected prototypes
    \item Additional visual results - domain-aligned prototypes
    
\end{itemize}

\subsection{Dataset Description}
\begin{figure}[h]
    \centering
    \includegraphics[width=0.35\textwidth]{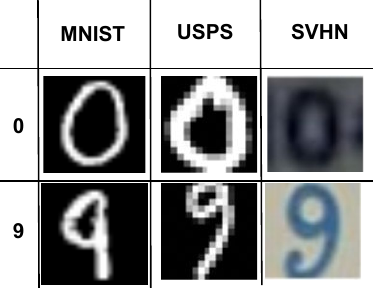}
    \caption{Some visual examples of the same digit classes (top: 0, bottom: 9) on the digit classification datasets - MNIST, USPS and SVHN. All samples were sampled from the train sets of each dataset.}
    \label{fig:digit-eg}
\end{figure}

\begin{figure}[h]
    \centering
    \includegraphics[width=0.4\textwidth]{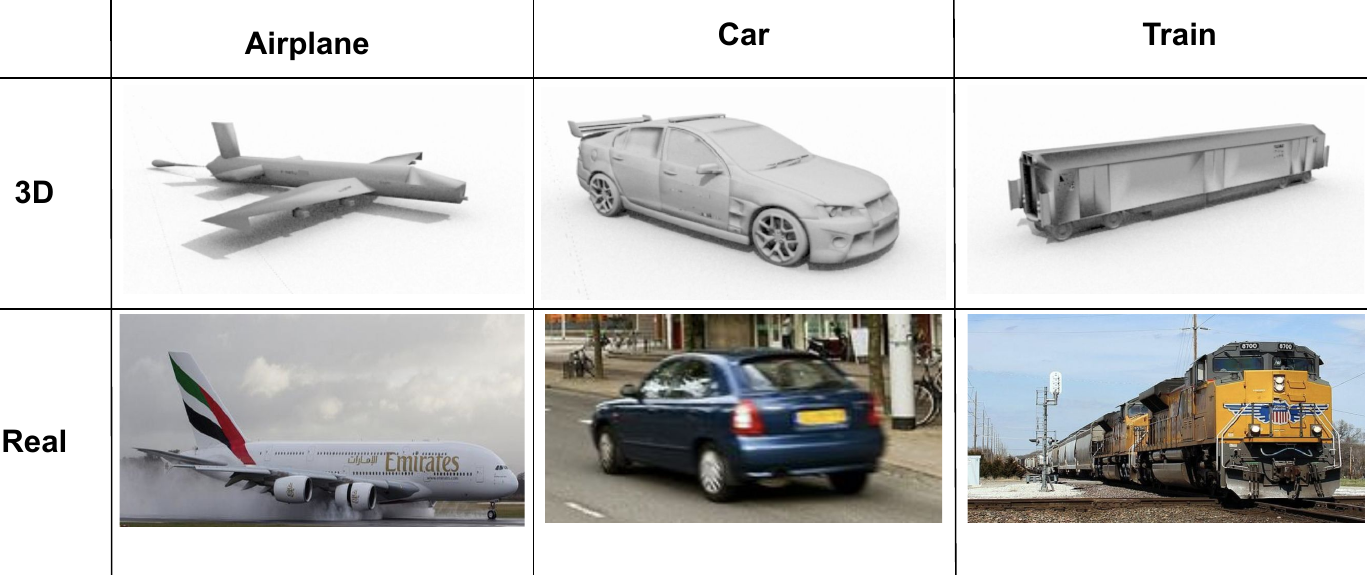}
    \caption{Some visual examples from the VisDA dataset corresponding to three classes - airplane, car and train. The top row demonstrates the training set of computer-rendered 3D images while the bottom row includes three examples of real images from the same classes.}
    \label{fig:visda-eg}
\end{figure}

\begin{figure}[h]
    \centering
    \includegraphics[width=0.35\textwidth]{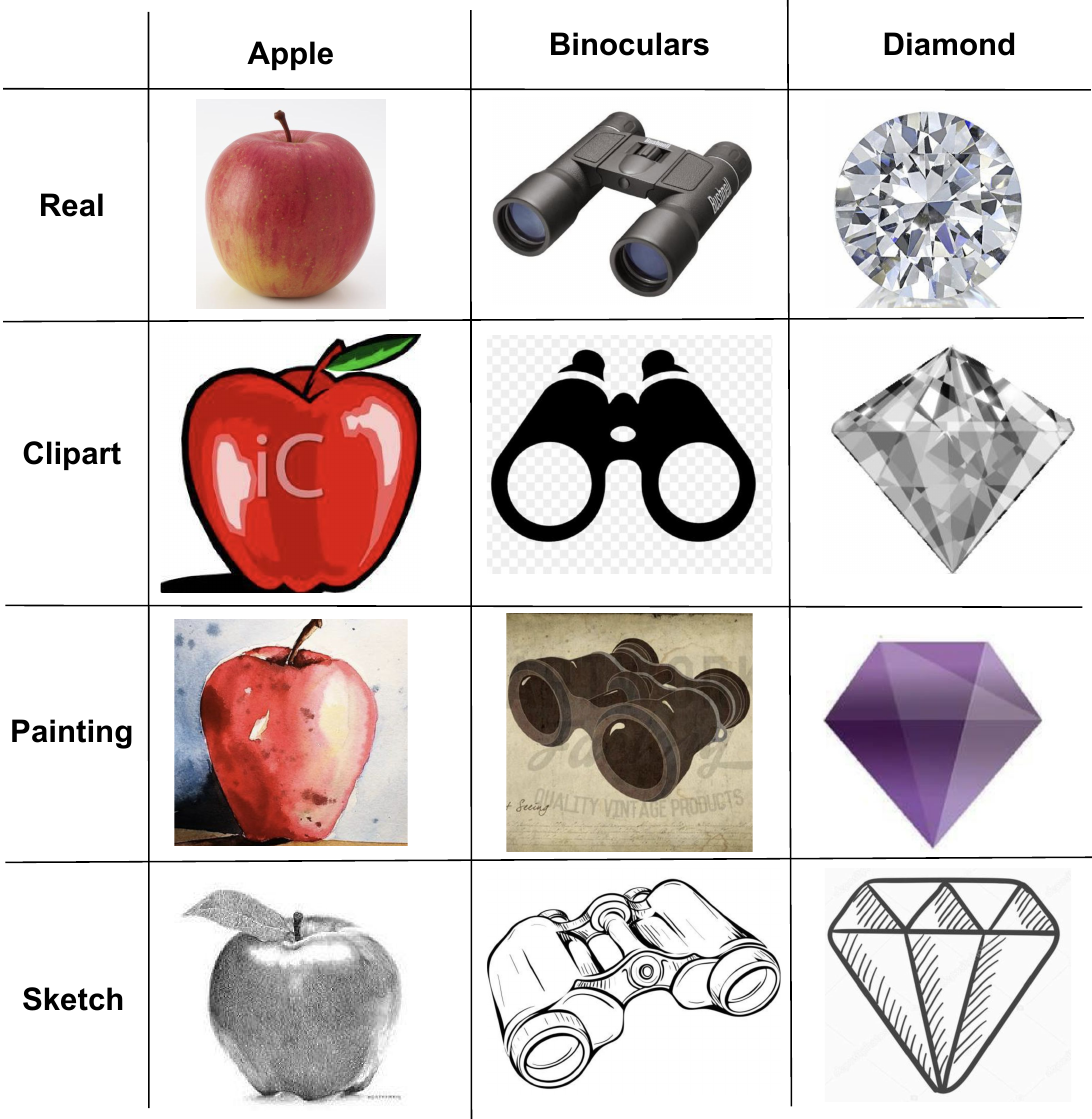}
    \vskip -10pt
    \caption{Some visual examples from the DomainNet dataset corresponding to three classes - apple, binoculars and diamond. The top row demonstrates images sampled from the Real (R), Clipart (C), Painting (P) and Sketch (S) domains.}
    \label{fig:domainnet-eg}
    \includegraphics[width=0.35\textwidth]{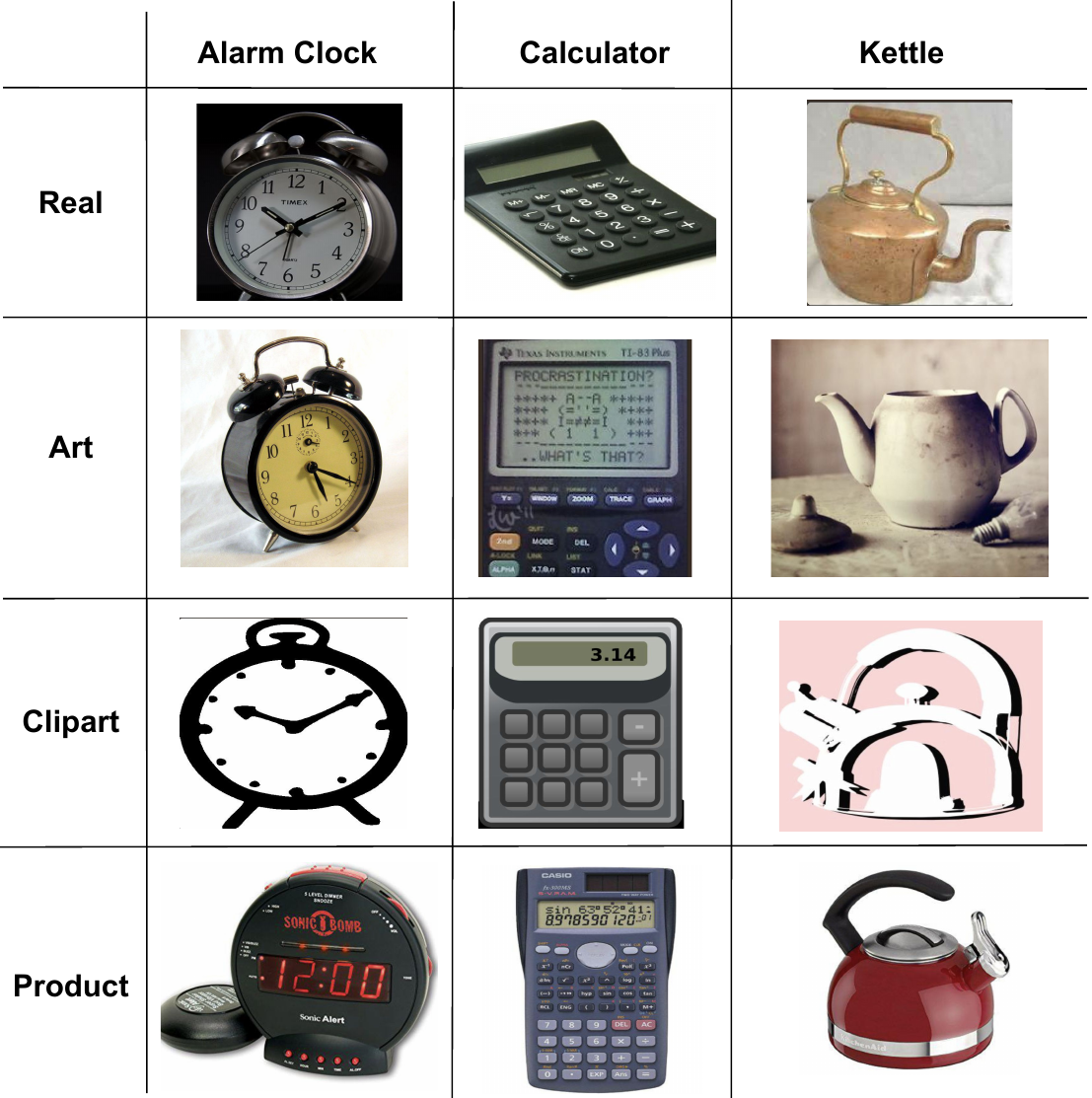}
    \vskip -10pt
    \caption{Some visual examples from the OfficeHome dataset corresponding to three classes - Alarm Clock, Calculator and Kettle. The rows demonstrate sample images from Real (R), Art (A), Clipart (C) and Product (P).}
    \label{fig:officehome-eg}
    \vskip -10pt
\end{figure}

A few examples from the training set of the datasets utilized in our approach are shown in Figures~\ref{fig:digit-eg} (Digits) for both tasks are shown in Figures~\ref{fig:digit-eg}, \ref{fig:visda-eg} (VisDA), \ref{fig:domainnet-eg} (DomainNet) and \ref{fig:officehome-eg} (OfficeHome).

\subsection{Training Procedure - Details}
Algorithm~\ref{algo:algo} depicts the overall pseudocode to train the Representative Concept Extraction (RCE) framework with Contrastive (CCL) and Prototype-based Grounding (PCG) regularization. The finer details of each part are listed as follows:

\noindent \textbf{RCE:} For networks $\mathbf{F}$ and $\mathbf{H}$, we utilize a Resnet34 architecture and initialize them with pre-trained Imagenet1k weights. For the network $\mathbf{A}$, we first utilize the element-wise vector product between the outputs of $\mathbf{F}$ and $\mathbf{H}$ and then pass the outputs through network $\mathbf{A}$, which is a shallow 2-layer fully connected network. For network $\mathbf{T}$, we utilize a 3-layer fully connected network which outputs a prediction with \textit{necessary and sufficient} concepts. The final prediction is a weighted sum of outputs of networks $\mathbf{A}$ and $\mathbf{T}$ followed by a softmax layer. The prediction loss is the standard cross-entropy loss.    

\noindent \textbf{PCG:} For selection of prototypical samples we utilize a combination of selection of source and target domains. We select 5 and 1 samples from the source and target domains respectively. Note that the PCG regularization starts after the 1st step. The grounding ensures that the outlying concept representations in the target domains are grounded to the abundant source domain representations. For concept bank, we utilize these 6 (5+1) prototypes for each class.

\begin{algorithm}[h]
\SetAlgoLined
\DontPrintSemicolon
\caption{Framework to train RCE using PCG+CCL}\label{algo:algo}

\SetKwInOut{Input}{Input}
\Input{Images: $X$, Labels: $Y$, Training Corpus: $\{(x_i, y_i) | x_i \in X, y_i \in Y\}$, Steps: $n$, Models: $F, G, H, A, T$,
Model Parameters: $\theta_F, \theta_G, \theta_H, \theta_A, \theta_T$,
Learning Rate: $\omega$, Positive Transformation Set: $T^+$,
Negative Transformation Set: $T^-$, Number of classes: $N$,
Batch size: $B$, Prototypes for source $S^s$, Prototypes for target $S^t$,
Concept Bank $\mathcal{C}$
}
\SetKwInOut{Output}{Output}
\Output{Trained models $F, G, H, A, T$}

$step \gets 0$\;
\For{$step \gets 1$ \KwTo $n$}
{
    \text{Sample a batch of training samples}\\
    $X, Y \gets \{(x_i, y_i) | i = 1, 2, \ldots, B\}$\;
    
    \For{$(x_i, y_i) \in \{(X, Y)\}$}
    {
        $\hat{y} \gets \omega_1 \cdot A(F(x_i) \odot H(x_i)) + \omega_2 \cdot T(F(x_i))$\;
        
        \text{Compute reconstruction and cross-entropy losses}
        $\mathcal{L}_{rec} \gets L(x_i, G(F(x_i))) + \lambda \cdot \|F(x_i)\|_1$\;
        
        $\mathcal{L}_{CE} \gets \mathcal{L}_{rec} + L(y, \hat{y})$\;
        
        \text{Compute positive and negative representations}
        $E^+ \gets \{F(t_j(x_i)) | t_j \in T^+\}$\; 
        $E^- \gets \{F(t_j(x_i)) | t_j \in T^-\}$\;
        
        \text{Compute SSL loss}\\
        $\mathcal{L}_{SSL} \gets -\log \left(\frac{\exp(s(e,e^+)/\tau)}{\sum_{e^- \in E^-}\exp(s(e,e^-)/\tau)}\right)$\;
        
        \text{Update concept bank}
        $\mathcal{C}_c \gets \frac{\mu}{|S^s_c|} \sum_{x \in S^s_c} F(x) + \frac{1-\mu}{|S^t_c|} \sum_{x \in S^t_c} F(x)$\;
        
        \text{Compute regularization losses}
        $\mathcal{L}_{grnd} \gets L(F(x_i), \mathcal{C})$\;
        $\mathcal{L}_{fid} \gets s(F(x_i), F(x_j)) \quad \text{for } y_i = y_j \in B$\;
        
        \text{Aggregate regularization losses}
        $\mathcal{L}_{CL} \gets \mathcal{L}_{SSL} + \lambda_1 \cdot \mathcal{L}_{grnd} + \lambda_2 \cdot \mathcal{L}_{fid}$\;
        
        \text{Compute total loss}
        $\mathcal{L} \gets \mathcal{L}_{CE} + \beta \cdot \mathcal{L}_{CL}$\;
        
        \text{Update model parameters}
        $\theta_{F, G, H, A, T} \gets \theta_{F, G, H, A, T} - \omega \nabla_{\theta_{F, G, H, A, T}} \mathcal{L}$\;
    }
}
\end{algorithm}

\subsection{Results on Key Hyperparameters}

\subsubsection{Number of Concepts}
The first 3 columns of Table~\ref{tab:officehome-ablation} list the domain adaptation performance on the OfficeHome dataset across 12 different data settings (listed in rows). We evaluate the performance by varying the number of concepts $C$ (and by extension, the relevance scores $S$).  We choose the base setting of the number of concepts being equal to the number of classes because we want each class to be represented by at least one concept.  We observe that increasing the number of concepts has no significant effect on the performance. This observation points to the fact that the relevant concept information is encoded in a few number of concepts. In other words, the concept vector is \textit{sparse}.

\subsubsection{Concept Dimensionality}
The last 3 columns of Table~\ref{tab:officehome-ablation} list performance by varying the concept dimensionality $d$ (dim). Note that non-unit dimensional concepts are not directly interpretable, and remain an active area of research \cite{sarkar2022framework}. Nevertheless, we report the performance numbers by varying the concept dimensionality. We observe that the with increasing concept dimensionality, the performance on target domains increases in almost all settings. This observation is expected for the following two reasons - 1) increasing concept dimensionality increases the richness of information encoded in each concept during contrastive learning and 2) increased dimensionality increases the expressiveness of the architecture itself.

\begin{table}[h]
    \centering
    \resizebox{0.49\textwidth}{!}{\begin{tabular}{c | c | ccc | ccc}
        \hline
        \textbf{Setting} & \textbf{S+T} & \multicolumn{6}{c}{\textbf{RCE+PCG+CCL}} \\
        \hline
         & & \multicolumn{3}{c|}{\textbf{Number of concepts}} & \multicolumn{3}{|c}{\textbf{Concept Dimension}} \\
        \hline
        & &  65 & 256 & 512 & 1 & 5* & 10* \\
        \hline
         A $\rightarrow$ C  & 54.0 & \textbf{58.7} & 58.7 & 58.7 & 58.7 & 58.8 & \textbf{59.0} \\
         A $\rightarrow$ P  & 73.1 & \textbf{73.7} & 73.7 & 73.7 & \textbf{73.7} & 73.7 & 73.7 \\ 
         A $\rightarrow$ R  & 74.2 & \textbf{75.0} & 75.0 & 75.0 & 75.0 & 75.1 & \textbf{75.4} \\
         C $\rightarrow$ A  & 57.6 & \textbf{58.0} & 58.0 & 58.0 & 58.0 & 58.0 & \textbf{58.1} \\
         C $\rightarrow$ P  & 72.3 & \textbf{71.9} & 71.9 & 71.9 & 71.9 & 71.9 & \textbf{72.0} \\
         C $\rightarrow$ R  & 68.3 & \textbf{68.9} & 68.9 & 68.9 & 68.9 & 68.9 & \textbf{68.9} \\
         P $\rightarrow$ A  & 63.5 & \textbf{62.1} & 62.1 & 62.1 & 62.1 & \textbf{62.1} & 62.0 \\
         P $\rightarrow$ C  & 53.8 & \textbf{55.4} & 55.4 & 55.4 & 55.4 & 55.4 & \textbf{55.5} \\
         P $\rightarrow$ R  & 73.1 & \textbf{74.8} & 74.8 & 74.8 & 74.8 & 74.6 & \textbf{75.0} \\
         R $\rightarrow$ A  & 67.8 & \textbf{67.2} & 67.2 & 67.2 & \textbf{67.2} & 67.2 & 67.0 \\
         R $\rightarrow$ C  & 55.7 & \textbf{60.2} & 60.2 & 60.2 & 60.2 & 60.2 & \textbf{60.3} \\
         R $\rightarrow$ P  & 80.8 & \textbf{81.3} & 81.3 & 81.3 & 81.3 & 81.3 & \textbf{81.5} \\
         \hline
    \end{tabular}}
    \caption{Effect of the most important hyperparameters - number of concepts (LEFT) and the dimensionality of concepts [RIGHT] on the domain adaptation performance. The asterisk (*) shows that non-unit concept dimensionality are not directly interpretable.} 
    \label{tab:officehome-ablation}
\end{table}

\subsubsection{Size of Representative set for PCG}
Table~\ref{tab:officehome-proto} shows the performance on the OfficeHome dataset for two settings of the pre-selected representative prototypes. For all experiments in the main paper, we utilize 5 prototypes from the source domain and 1 from the target domain - for a total of 6 prototype samples for grounding. Note that it is not usually possible to use a lot of prototypes from the target domain as our setting corresponds to the 3-shot setting in \cite{yu2023semi}. We show the performance on 5 and 7 selected prototypes on the source domain in Table~\ref{tab:officehome-proto}. We observe that increasing the number of prototypes does not result in an improvement of performance, in fact performance does not significantly change and in a few cases, performance drops. This observation implies that only a minimum number of necessary and sufficient grounding set of prototypes is required. This observation is consistent with intuition because if more than a requisite prototypes are selected, the computation time for concept representations will increase.

\subsubsection{Distances from the Concept Representation prototypes}
Table~\ref{tab:officehome-proto} lists the average normalized distance of the concept representations of the target domain from the concept representations associated with the selected prototypes with varying values of $\lambda_1$ which controls the effect of supervision in PCG. We choose 3 different values for demonstrating the ablation results - $\lambda_1=0$ for no regularization and $\lambda_1=1$ for very high regularization. We observe that both cases lack generalization performance implying a tradeoff between regularization and generalization.

\begin{table*}[t]
    \centering
    \resizebox{0.7\textwidth}{!}{\begin{tabular}{c | c | cc|cc|cc | cc|cc|cc}
        \hline
        \textbf{Setting} & \textbf{S+T} & \multicolumn{12}{c}{\textbf{RCE+PCG+CCL}} \\
        \hline
         \multicolumn{2}{c|}{\textbf{\# of prototypes}} & \multicolumn{6}{c|}{\textbf{Source Prototypes = 5}} & \multicolumn{6}{|c}{\textbf{Source Prototypes = 7}} \\
        \hline
        & & \multicolumn{2}{c}{$\lambda_1$=0.1} & \multicolumn{2}{c}{$\lambda_1$=0} & \multicolumn{2}{c|}{$\lambda_1$=1}  & \multicolumn{2}{c}{$\lambda_1$=0.1} & \multicolumn{2}{c}{$\lambda_1$=0} &\multicolumn{2}{c}{$\lambda_1$=1}   \\
        \hline
        & & Dist & Perf & Dist & Perf & Dist & Perf & Dist & Perf & Dist & Perf & Dist & Perf  \\
        \hline
         A $\rightarrow$ C  & 54.0 & 0.32 & \textbf{58.7} & 1.8  & 53.3 & 0.01 & 54.0 & 0.32 & \textbf{58.8} & 1.8 & 53.3 & 0.01 & 53.1 \\
         A $\rightarrow$ P  & 73.1 & 0.30 & \textbf{73.7} & 1.7  & 74.1 & 0.01 & 69.2 & 0.31 & \textbf{73.7} & 1.7 & 74.1 & 0.01 & 68.4 \\ 
         A $\rightarrow$ R  & 74.2 & 0.30 & \textbf{75.0} & 1.8  & 74.8 & 0.01 & 73.1 & 0.31 & \textbf{74.9} & 1.8 & 74.8 & 0.01 & 72.3 \\
         C $\rightarrow$ A  & 57.6 & 0.27 & \textbf{58.0} & 1.1  & 57.1 & 0.02 & 55.1 & 0.24 & \textbf{58.1} & 1.1 & 57.1 & 0.01 & 52.6 \\
         C $\rightarrow$ P  & 72.3 & 0.30 & \textbf{71.9} & 1.2  & 70.2 & 0.02 & 70.1 & 0.30 & \textbf{71.3} & 1.2 & 70.2 & 0.01 & 68.1\\
         C $\rightarrow$ R  & 68.3 & 0.24 & \textbf{68.9} & 1.5  & 67.8 & 0.01 & 66.4 & 0.24 & \textbf{68.1} & 1.5 & 67.8 & 0.01 & 64.8\\
         P $\rightarrow$ A  & 63.5 & 0.21 & \textbf{63.5} & 1.4  & 62.1 & 0.01 & 61.2 & 0.21 & \textbf{63.6} & 1.4 & 62.1 & 0.01 & 60.8 \\
         P $\rightarrow$ C  & 53.8 & 0.22 & \textbf{55.4} & 1.4  & 54.2 & 0.01 & 53.6 & 0.22 & \textbf{55.4} & 1.4 & 54.2 & 0.01 & 52.1\\
         P $\rightarrow$ R  & 73.1 & 0.26 & \textbf{74.8} & 1.8  & 73.3 & 0.01 & 70.1 & 0.26 & \textbf{74.7} & 1.8 & 73.3 & 0.01 & 68.6\\
         R $\rightarrow$ A  & 67.8 & 0.18 & \textbf{67.2} & 0.91 & 67.0 & 0.01 & 66.9 & 0.18 & \textbf{67.2} & 0.91 & 67.0 & 0.01 & 65.5 \\
         R $\rightarrow$ C  & 55.7 & 0.18 & \textbf{60.2} & 0.93 & 58.3 & 0.01 & 55.1 & 0.18 & \textbf{60.3} & 0.93 & 58.3 & 0.01 & 53.9 \\
         R $\rightarrow$ P  & 80.8 & 0.19 & \textbf{81.3} & 0.94 & 80.8 & 0.01 & 80.0 & 0.19 & \textbf{81.3} & 0.94 & 80.8 & 0.01 & 78.9\\
         \hline
    \end{tabular}}
    \caption{Effect of the most important hyperparameters - size of the selected prototypes - sizes 5 and 7. The columns \textbf{Dist} refer to the average normalized L2 distance between the concept representations and concept prototype representations in the concept space while \textbf{Perf} refers to the performance. Note that there is a balance between the distance between prototypes and representations - too low distances ($\lambda_1=1$) fail to generalize to the target domain effectively while no distance regularization ($\lambda_1=0$) performs very close to unregularized approaches implying the need for PCG. } 
    \label{tab:officehome-proto}
    \begin{tabular}{c | c|c|c | c|c | c|c|c|c |  c|c|c|c}
    \hline
     & \multicolumn{3}{c|}{\textbf{Digit}} & \multicolumn{2}{c|}{\textbf{VisDA}}  & \multicolumn{4}{c|}{\textbf{DomainNet}} & \multicolumn{4}{c}{\textbf{OfficeHome}}\\
    \hline
         & M &  U &  S & 3D &  R &  A & C & P & R & C & P & R & S\\
         \hline
          SENN & 0.81 &  0.74 & 0.61  & 0.43 & 0.48  & 0.21 & 0.24 & 0.26 & 0.30 & 0.31 & 0.27 & 0.29 & 0.30  \\
         
          DSENN &  0.79 &  0.71 & 0.63  & 0.42 & 0.46  & 0.14 & 0,22 & 0.21 & 0.27 & 0.29 & 0.23 & 0.29 & 0.32  \\
          
          BotCL & 0.93 & \textbf{0.94}  & 0.89  & 0.79 &  \textbf{0.83} & \textbf{0.49} & 0.55 & 0.51 & 0.58 & \textbf{0.73} & 0.66  & 0.61 & 0.64 \\
          
          UnsupervisedCBM & 0.79 &  0.74 &  0.63 & 0.42 & 0.46  & 0.21 & 0.24 & 0.26 & 0.30   & 0.31 & 0.27 & 0.29 & 0.30  \\
          \hline
          
          RCE & 0.86 &  0.80 & 0.73  & 0.72 & 0.74  & 0.39 & 0.50 & 0.45 & 0.42 & 0.54 & 0.49 & 0.50 & 0.49 \\
          
          RCE+PCG &  0.94 &  0.94 & 0.89  & 0.81 & 0.81  & 0.47 & 0.56 & 0.51 & 0.59 & 0.70 & 0.67 & 0.61 & 0.63 \\
          
          \textbf{RCE+PCG+CCL} &  \textbf{0.94} &  0.94 & \textbf{0.89}  & \textbf{0.81} & 0.81  & 0.47 & \textbf{0.55} & \textbf{0.52} & \textbf{0.59} & 0.71 & \textbf{0.68} & \textbf{0.63} & \textbf{0.64} \\
          \hline
    \end{tabular}%
     \caption{Average Intra-class Concept Fidelity scores for each domain for all settings where the domain is the target. Rows S, D, B and U respectively correspond to \underline{S}ENN, \underline{D}iSENN, \underline{B}otCL and \underline{U}nsupCBM. Similarly, R, P and C correspond to RCE, RCE+PCG and RCE+PCG+CCL . The columns show the domains in each dataset.}
    \label{tab:jsi-appendix} 
\end{table*}

\subsection{Concept Fidelity Analysis}
Table~\ref{tab:jsi-appendix} lists the consolidated concept fidelity scores of all four datasets. Note: This table is a complete version of the Table~\ref{tab:jsi} in the main text. We see that the concept overlap on all datasets is highest in either our approach or BotCL, both approaches with explicit fidelity regularizations. This demonstrates the efficacy of our approach in maintaining concept fidelity.

\subsection{Baseline Replication}
We compare our approach against 4 baselines - SENN \cite{alvarez2018towards}, DiSENN \cite{disenn}, BotCL\cite{wang2023learning} and UnsupervisedCBM\cite{sawada2022concept}. Even though none of the approaches incorporate domain adaptation as an evaluation method, we utilize the proposed methodology directly in our settings. Proper care has been taken to ensure high overlap with the intended use and carefully introduced modifications proposed by us, listed below: 
\begin{itemize}
    \item \textbf{SENN and DiSENN:} We utilize the well tested publicly available code\footnote{\url{https://github.com/AmanDaVinci/SENN}} as the basic framework. We modify SENN and DiSENN to include Resnet34 (for objects) and the decoder is kept the same for all setups discussed. Specifically, due to the very slow computation of the robustness loss $\mathcal{L}_h$ on bigger networks like Resnet34, we only compute it once every 10 steps.
    \item \textbf{BotCL:} We utilize the publicly available code \footnote{\url{https://github.com/wbw520/BotCL}}. We utilize the same network architecture - LeNet for digits and Resnet34 for objects. Additionally, we amend the LeNet architecture to fit BotCL framework.
    \item \textbf{UnsupervisedCBM:} Unsupervised CBM is hard to train as it contains a mixture of supervised and unsupervised concepts. However, our approach does not utilize supervision, so we only consider the unsupervised concepts and replace the supervised concepts with the one-hot encoding of the classes of the images. We utilize a fully connected layer for the discriminator network while simultaneously training a decoder. Though a publicly available version of UnsupCBM is not available, we are successful in the replication of its main results.
\end{itemize}

\newpage
\clearpage
\begin{figure*}[h!]
    \centering
    \begin{subfigure}[b]{0.75\textwidth}
        \includegraphics[width=\textwidth]{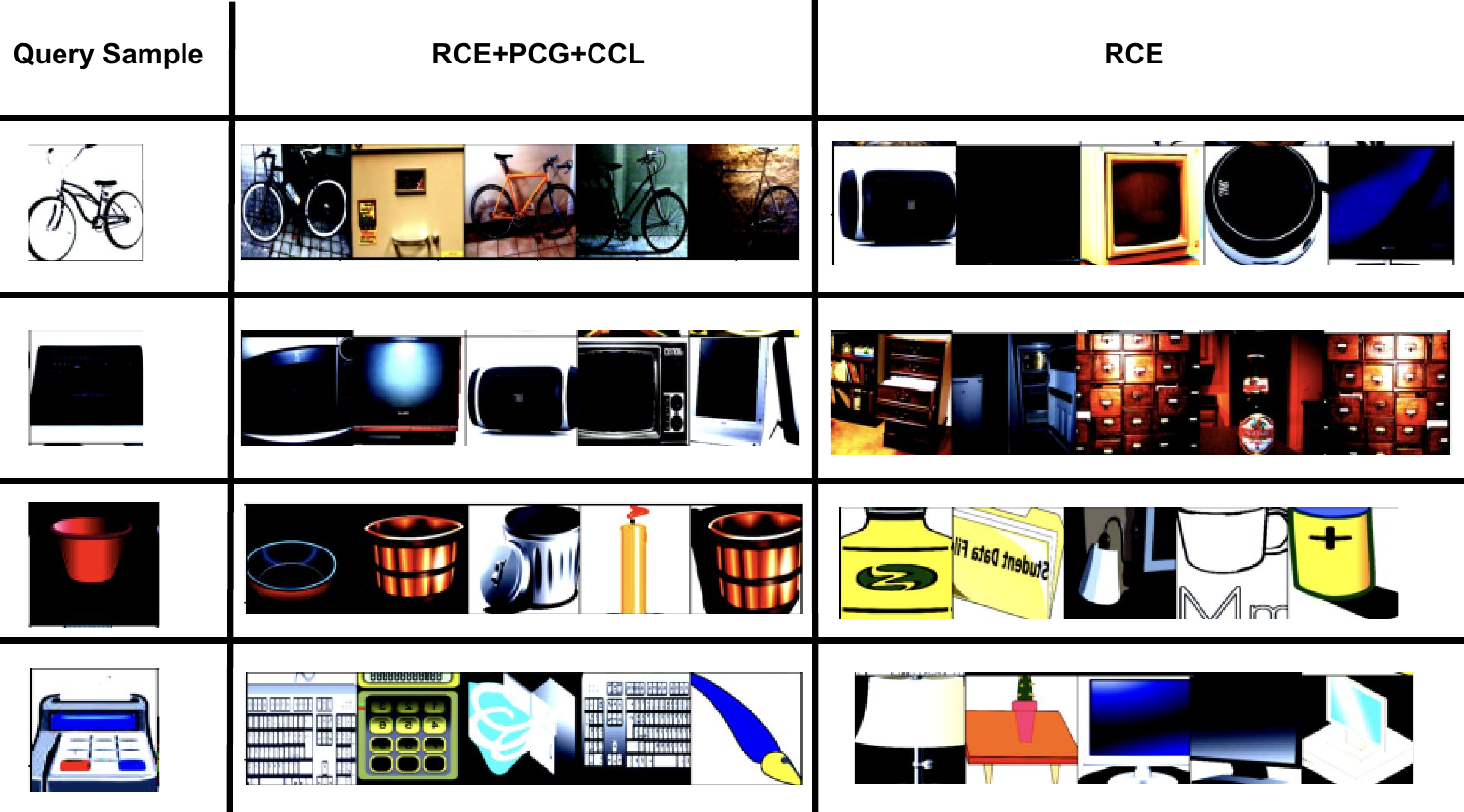}
        \caption{Top-5 most important prototypes associated with the most important concept associated with the query sample - on a model trained using our methodology on the OfficeHome dataset for the (from top) Real, Product, Art and Clipart domains respectively where the domains are target. The prototypes on the left are predicted by RCE models trained using PCG+CCL while the ones on the left are predicted by models trained without PCG+CCL.}
        \label{fig:examples-alignment-query}
    \end{subfigure}
    
    \begin{subfigure}[b]{0.75\textwidth}
        \includegraphics[width=\textwidth]{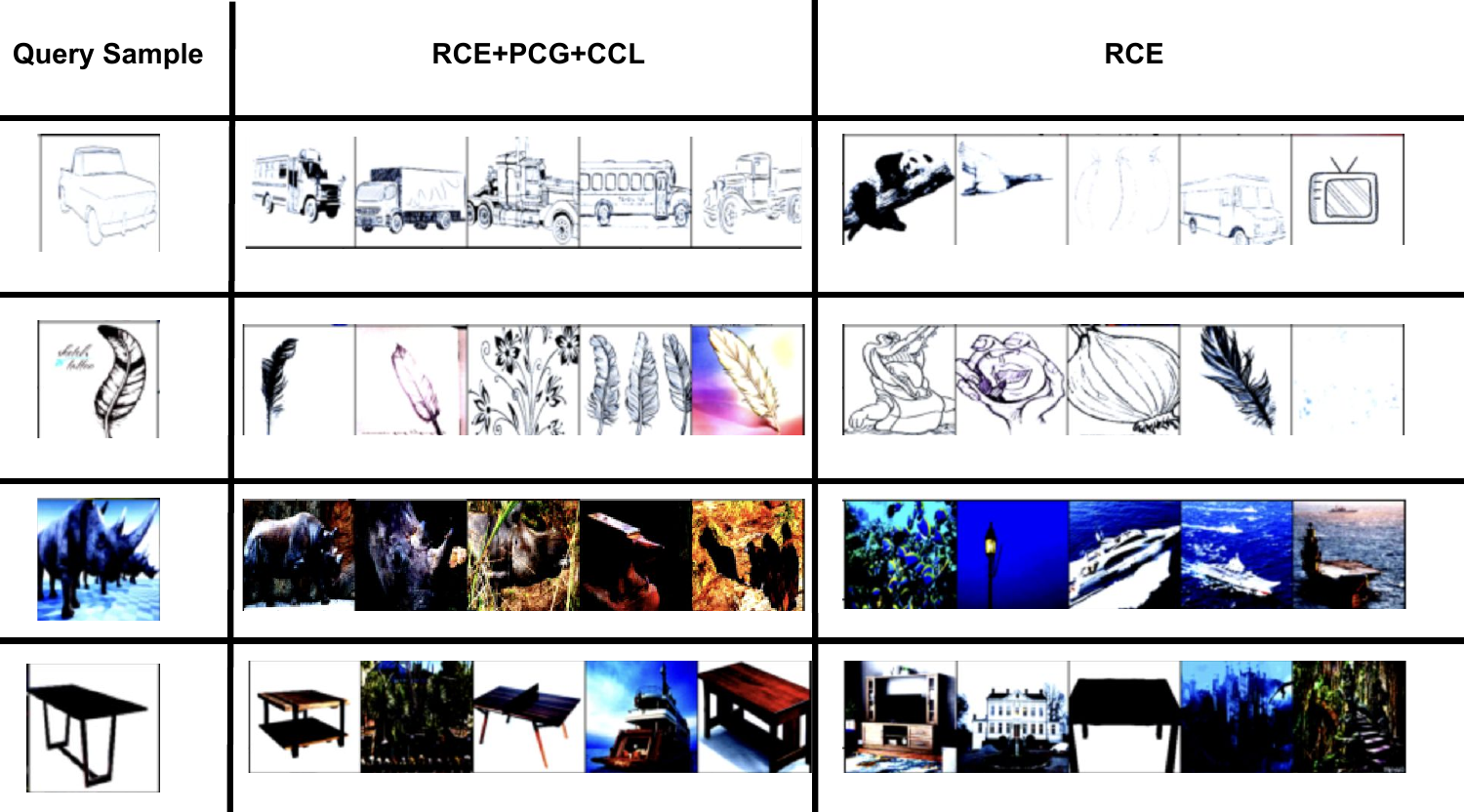} 
        \caption{Top-5 most important prototypes associated with the most important concept associated with the query sample - on a model trained using our methodology on the DomainNet dataset for the (from top) Sketch, Painting, and Real domains respectively where the domains are target. The prototypes on the left are predicted by RCE models trained using PCG+CCL while the ones on the left are predicted by models trained without PCG+CCL.}
    \label{fig:examples-alignment-query-domainnet}
    \end{subfigure}
    \caption{Selected prototypes for the (a) OfficeHome and (b) DomainNet datasets repectively.}
\end{figure*}

\subsection{Additional Visual Results - Selected Prototypes}
Figures~\ref{fig:examples-alignment-query} and ~\ref{fig:examples-alignment-query-domainnet} showcase the most top-5 most important prototypes concerning a given query image for the OfficeHome and DomainNet datasets respectively. For each row, the model settings are where the query and prototypes are in the target domain. We show results on all domains for both datasets, to demonstrate the efficacy of our proposed approach, which can generalize to all domains. Note that RCE is an overparameterized version of SENN, hence the performance with respect to baselines remains identical. Our proposed approach can explain each query image with relevant prototypes as opposed to the baselines where the prototypes are barely relevant.

\subsection{Additional Visual Results - Domain Aligned}
\begin{figure*}[h!]
    \centering
    \begin{subfigure}[b]{0.75\textwidth}
        \includegraphics[width=\textwidth]{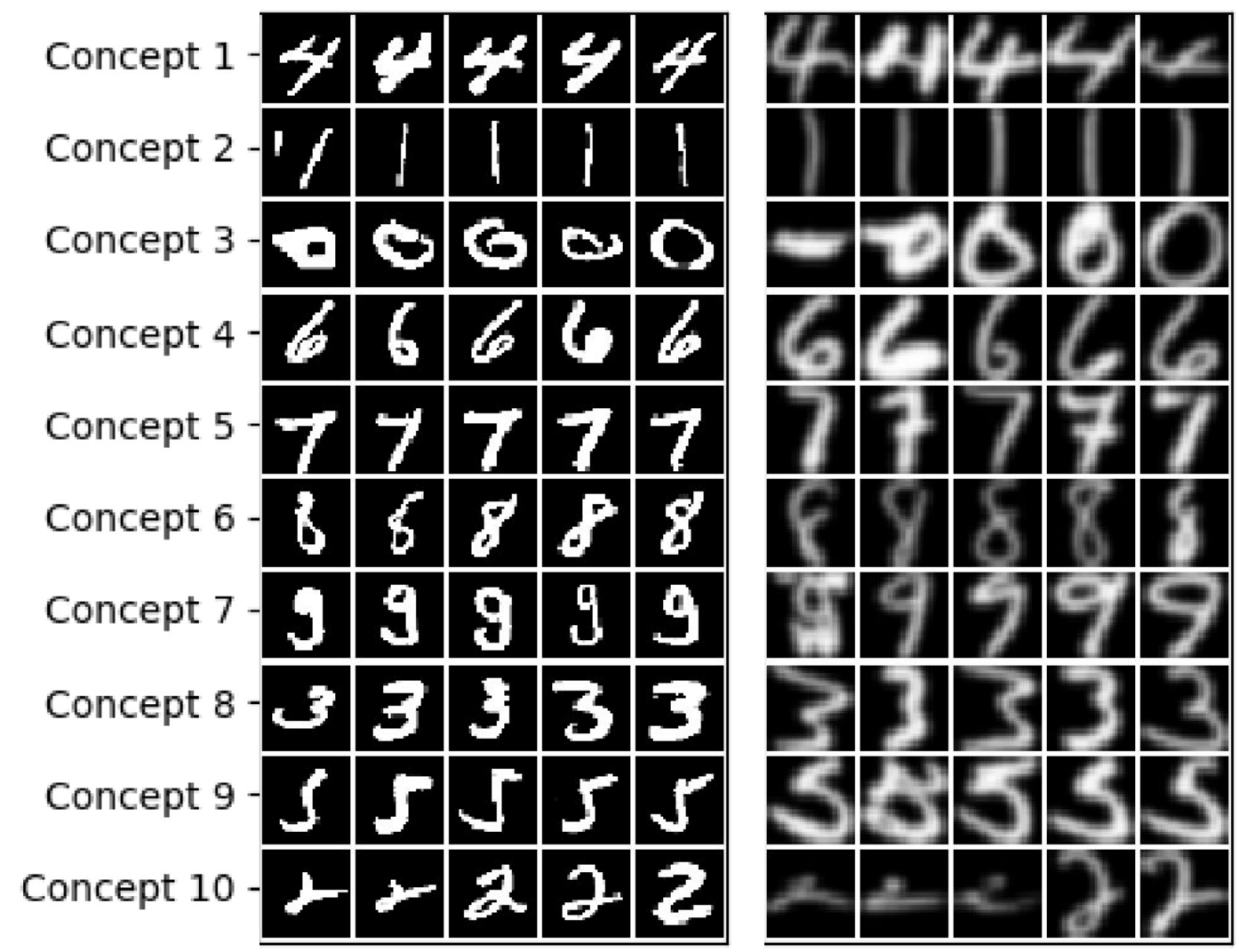}
        \caption{Prototypes that maximally activate each concept learned. As can be observed, there is high agreement between prototypes across all domains.}
        \label{fig:examples-alignment-appendix}
    \end{subfigure}
    
    \begin{subfigure}[b]{0.75\textwidth}
        \includegraphics[width=\textwidth]{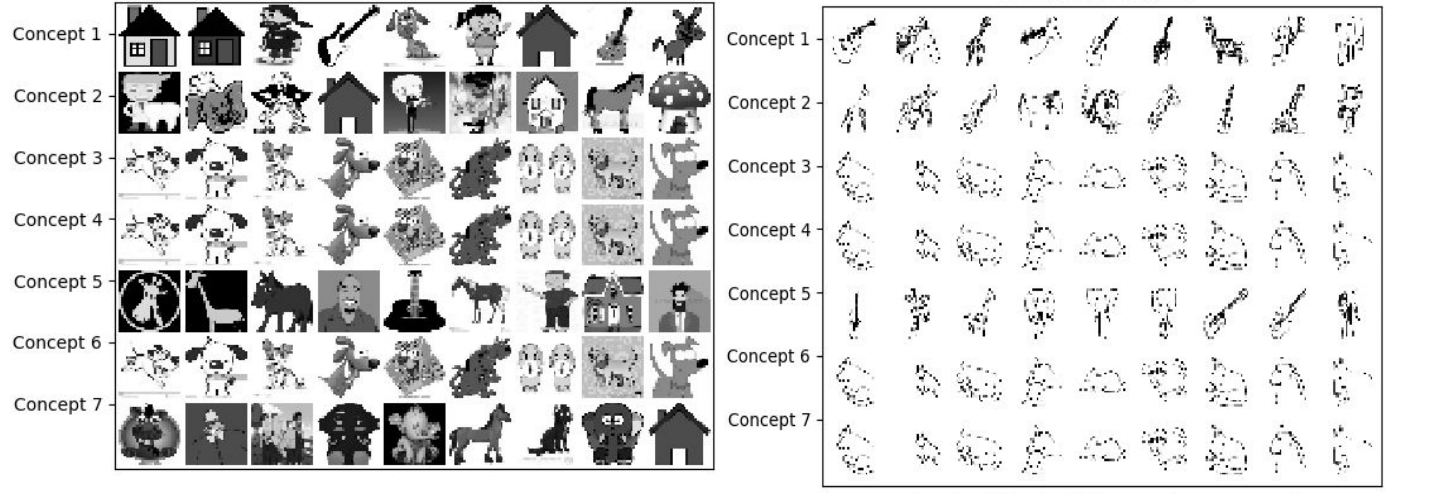} 
        \caption{Prototypes that maximally activate each concept learned. As can be observed, there is high agreement between prototypes across both art and sketch domains.}
    \label{fig:examples-alignment-appendix2}
    \end{subfigure}
    \caption{Domain aligned prototype selection for [TOP] Digits - MNIST and USPS and [BOTTOM] PACS dataset.}
\end{figure*}

Figures~\ref{fig:examples-alignment-appendix} and ~\ref{fig:examples-alignment-appendix2} showcase the most top-5 most important prototypes concerning Digit and VisDA datasets respectively. For each row, the model settings on the left show the prototypes in the source and on the right show target domain. Our proposed approach can explain each concept with relevant prototypes across domains.

